\documentclass{article}

\usepackage{microtype}
\usepackage{graphicx}
\usepackage{subfigure}
\usepackage{booktabs} % for professional tables
\usepackage{multirow}

\usepackage{hyperref}

\usepackage[accepted]{icml2024}

\usepackage{amsmath}
\usepackage{amssymb}
\usepackage{mathtools}
\usepackage{amsthm}
\usepackage{bm}

\usepackage[capitalize,noabbrev]{cleveref}

\theoremstyle{plain}
\newtheorem{theorem}{Theorem}[section]
\newtheorem{proposition}[theorem]{Proposition}

\theoremstyle{definition}

\theoremstyle{remark}

\usepackage[textsize=tiny]{todonotes}

\icmltitlerunning{Image Restoration Through Generalized Ornstein-Uhlenbeck Bridge}

\begin{document}

\twocolumn[
\icmltitle{Image Restoration Through Generalized Ornstein-Uhlenbeck Bridge}




\begin{icmlauthorlist}
\icmlauthor{Conghan Yue}{yyy}
\icmlauthor{Zhengwei Peng}{yyy}
\icmlauthor{Junlong Ma}{yyy}
\icmlauthor{Shiyan Du}{yyy}
\icmlauthor{Pengxu Wei}{yyy}
\icmlauthor{Dongyu Zhang}{yyy}

\end{icmlauthorlist}

\icmlaffiliation{yyy}{Department of Computer Science, Sun Yat-sen University, Guangzhou, Guangdong, China}

\icmlcorrespondingauthor{Conghan Yue}{yuech5@mail2.sysu.edu.cn}

\icmlkeywords{Diffusion Model, Diffusion Bridge, Image Restoration}

\vskip 0.3in
]



\printAffiliationsAndNotice{}  % leave blank if no need to mention equal contribution

\begin{abstract}
Diffusion models exhibit powerful generative capabilities enabling noise mapping to data via reverse stochastic differential equations. However, in image restoration, the focus is on the mapping relationship from low-quality to high-quality images. Regarding this issue, we introduce the Generalized Ornstein-Uhlenbeck Bridge (GOUB) model.
By leveraging the natural mean-reverting property of the generalized OU process and further eliminating the variance of its steady-state distribution through the Doob’s \textit{h}–transform, we achieve diffusion mappings from point to point enabling the recovery of high-quality images from low-quality ones. Moreover, we unravel the fundamental mathematical essence shared by various bridge models, all of which are special instances of GOUB and empirically demonstrate the optimality of our proposed models. Additionally, we present the corresponding Mean-ODE model adept at capturing both pixel-level details and structural perceptions. Experimental outcomes showcase the state-of-the-art performance achieved by both models across diverse tasks, including inpainting, deraining, and super-resolution.
Code is available at \url{https://github.com/Hammour-steak/GOUB}.
\end{abstract}

\section{Introduction}
Image restoration involves the restoring of high-quality (HQ) images from their low-quality (LQ) version \cite{banham1997digital,zhou1988image,liang2021swinir,luo2023refusion}, which is often characterized as an ill-posed inverse problem due to the loss of crucial information during the degradation from high-quality images to low-quality images. 
It encompasses a suite of classical tasks, including image deraining \cite{zhang2017convolutional,yang2020single,xiao2022image}, denoising \cite{zhang2018ffdnet,li2022all,soh2022variational,zhang2023mm}, deblurring \cite{yuan2007image,kong2023efficient}, inpainting \cite{jain2023keys,zhang2023towards}, and super-resolution \cite{dong2015image,zamfir2023towards,wei2023taylor}, among others.

Diffusion models \cite{sohl2015deep,ho2020denoising,song2019generative,song2020score,karras2022elucidating} have also been applied to image restoration, yielding favorable results \cite{ho2022classifier,wang2023unlimited,su2022dual,shi2023diffusion}.
They mainly follow the standard forward process, diffusing images to pure noise and using low-quality images as conditions to facilitate the generation process of high-quality images \cite{dhariwal2021diffusion, ho2022classifier, kawar2021snips,saharia2022image,kawar2022denoising,chung2022improving,chung2022diffusion,wang2023unlimited}. 
However, these approaches require the integration of substantial prior knowledge specific to each task such as degradation matrices, limiting their universality.

Furthermore, some studies have attempted to establish a point-to-point mapping from low-quality to high-quality images, learning the general degradation and restoration process and thus circumventing the need for additional prior information for modeling specific tasks \cite{chen2022simple,cui2023focal,lee2024ugpnet}.
In terms of diffusion models, this mapping can be realized through the bridge \cite{liu2022deep, su2022dual, liu20232}, a stochastic process with fixed starting and ending points. By assigning high-quality and low-quality images to the starting and ending points, and initiating with the low-quality images, high-quality images can be obtained by applying the reverse diffusion process, thereby enabling image restoration. 
However, some bridge models face challenges in learning likelihoods \cite{liu2022deep}, necessitating reliance on cumbersome iterative approximation methods \cite{de2021diffusion,su2022dual,shi2023diffusion}, which pose significant constraints in practical applications; others do not consider the selection of diffusion process and ignore the optimality of diffusion process \cite{liu20232,li2023bbdm,zhou2023denoising}, thus may introducing unnecessary costs and limiting the performance of the model.

\begin{figure*}[t]
  \centering
  \includegraphics[width=0.85\textwidth]{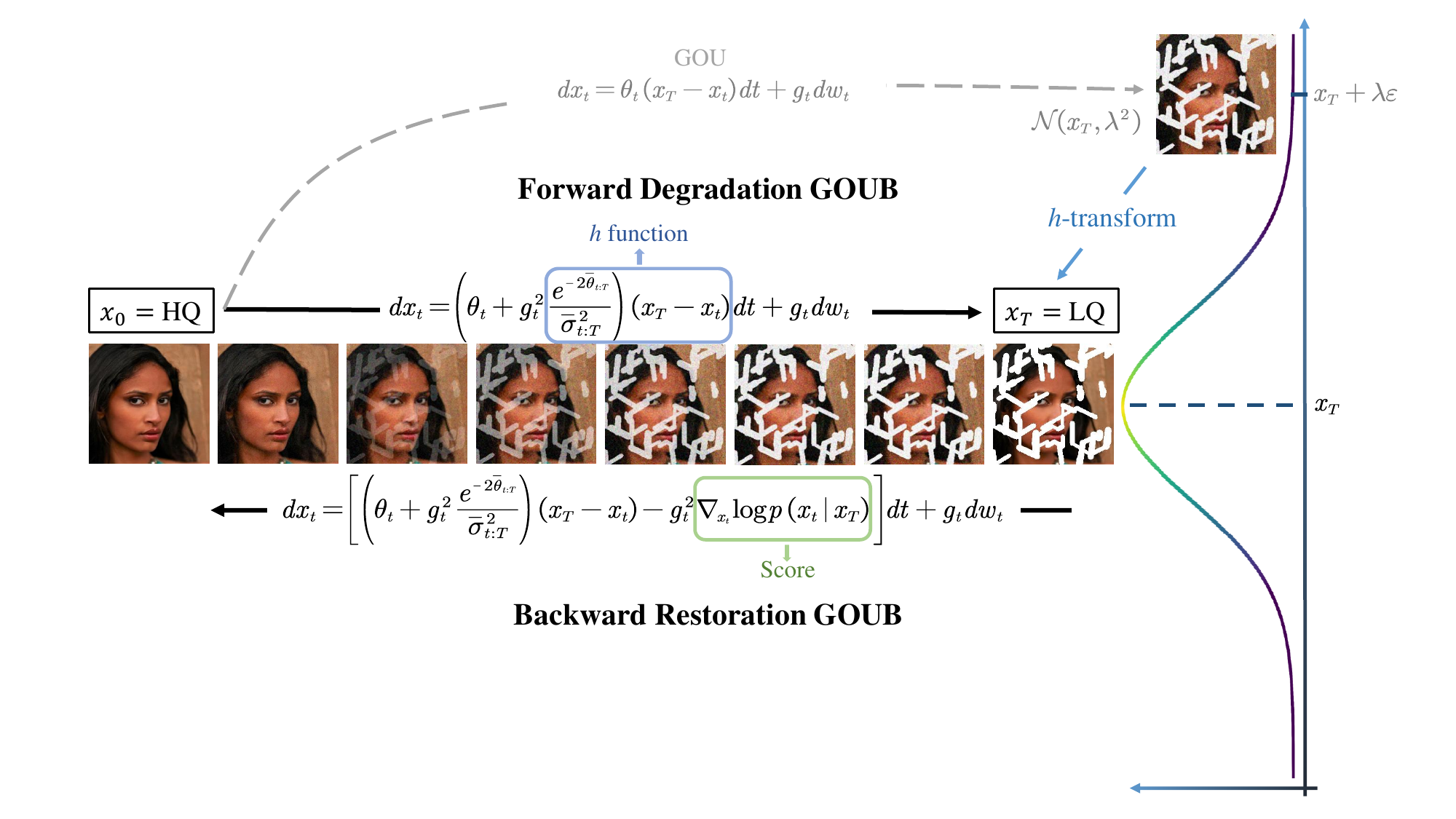}
  \caption{Overview of the proposed GOUB for image restoration. 
  The GOU process is capable of transferring an HQ image into a noisy LQ image. Additionally, through the application of \textit{h}-transform, we can eliminate the noise on LQ, enabling the GOUB model to precisely bridge the gap between HQ and LQ.}
  \label{framework}
  \vskip -0.1in
\end{figure*}

This paper proposed a novel image restoration bridge model, the Generalized Ornstein-Uhlenbeck Bridge (GOUB), depicted in Figure \ref{framework}. Owing to the mean-reverting properties of the Generalized Ornstein-Uhlenbeck (GOU) process, it gradually diffuses the HQ image into a noisy LQ state (denoted as $\mathbf x_T +\lambda\epsilon$ in Figure \ref{framework}). By applying Doob's \textit{h}-transform on GOU, we modify the diffusion process to eliminate noise on $\mathbf x_T$ to directly bridge the HQ image and its LQ counterpart. The model initiates a point-to-point forward diffusion process and learns its reverse through maximum likelihood estimation, thereby ensuring it can restore a low-quality image to the corresponding high-quality image avoiding the limitation of generality and costly iterative approximation.
Our main contributions can be summarized as follows:
\begin{itemize}
	\item We introduce a novel image restoration bridge model GOUB which eliminates variance of the ending point on the GOU process, directly connecting the high and low-quality images and is particularly expressive in deep visual features and diversity.
	\item Benefiting from the distinctive features of the parameterization mechanism, we introduce the corresponding Mean-ODE model, demonstrating a strong ability to capture pixel-level details and structural perceptions. 
    \item We uncover the mathematical essence of several bridge models, all of which are special cases of the GOUB, and empirically demonstrate the optimality of our proposed models.
	\item Our model has achieved state-of-the-art results on numerous image restoration tasks, such as inpainting, deraining, and super-resolution.
\end{itemize}

\section{Preliminaries}

\subsection{Score-based Diffusion Model}
The score-based diffusion model \cite{sohl2015deep, ho2020denoising, song2020score} is a category of generative model that seamlessly transitions data into noise via a diffusion process and generates samples by learning and adapting the reverse process \cite{anderson1982reverse}.
Assuming a dataset consists of $n$ dimensional independent identically distributed (i.i.d.) samples, following an unknown distribution denoted by $p(\mathbf{x_0})$. The time-dependent forward process of the diffusion model can be described by the following SDE: 
\begin{equation}\label{eq1}
\mathrm{d}\mathbf{x}_t=\mathbf{f}\left( \mathbf{x}_t,t \right) \mathrm{d}t+g_t \mathrm{d}\mathbf{w}_t, 
\end{equation}
where $\mathbf{f}:\mathbb{R}^n\rightarrow \mathbb{R}^n$ is the drift coefficient, $g_t:\mathbb{R}\rightarrow \mathbb{R}$ is the scalar diffusion coefficient and $\mathbf{w}_t$ denotes the standard Brownian motion. Typically, $p(\mathbf{x}_0)$ evolves over time $t$ from 0 to a sufficiently large $T$ into $p(\mathbf{x}_T)$ through the SDE, such that $p(\mathbf{x}_T)$ will approximate a standard Gaussian distribution $p_{\text{prior}}(\mathbf{x})$. Meanwhile, the forward SDE has a corresponding reverse time SDE \cite{anderson1982reverse} whose closed form is given by:
\begin{equation}
\label{eq2}
\mathrm{d}\mathbf{x}_t=\left[\mathbf{f}\left( \mathbf{x}_t,t \right)-g^2_t\nabla_{\mathbf{x}_t}\log p(\mathbf{x}_t)\right] \mathrm{d}t+g_t \mathrm{d}\mathbf{w}_t.
\end{equation}
Starting from time $T$, $p(\mathbf{x}_T)$ can progressively transform to $p(\mathbf{x}_0)$ by traversing the trajectory of the reverse SDE.
The score $\nabla_{\mathbf{x}_t}\log p(\mathbf{x}_t)$ can generally be parameterized as $\mathbf{s}_{\bm\theta}(\mathbf{x}_t, t)$ and employ conditional score matching \cite{vincent2011connection} as the loss function for training:
\begin{equation}\label{eq3}
\scriptsize
\begin{aligned}
\mathcal{L}
&=\frac{1}{2}\int_0^T\mathbb{E} _{ \mathbf{x}_t }\Bigg[ \lambda \left( t \right) \left\| \nabla _{\mathbf{x}_t}\log p\left( \mathbf{x}_t \right) -\mathbf{s}_{\bm\theta}\left( \mathbf{x}_t,t \right) \right\|^2 \Bigg] \mathrm{d}t \\
&\propto\frac{1}{2}\int_0^T\mathbb{E} _{ \mathbf{x}_0, \mathbf{x}_t }\Bigg[ \lambda \left( t \right) \left\| \nabla _{\mathbf{x}_t}\log p\left( \mathbf{x}_t\mid \mathbf{x}_0 \right) -\mathbf{s}_{\bm\theta}\left( \mathbf{x}_t,t \right) \right\|^2 \Bigg] \mathrm{d}t,
\end{aligned}
\end{equation}
where $\lambda(t)$ serves as a weighting function, and if selected as $g^2_t$ that yields a more optimal upper bound on the negative log-likelihood \cite{song2021maximum}. The second line is actually the most commonly used, as the conditional probability $p(\mathbf{x}_t \mid \mathbf{x}_0)$ is generally accessible. Ultimately, one can sample $\mathbf{x}_T$ from the prior distribution $p(\mathbf{x}_T) \approx p_{\text{prior}}(\mathbf{x})$ and obtain the $\mathbf{x}_0$ through the numerical solution of Equation \eqref{eq2} via iterative steps, thereby completing the generation process.

\subsection{Generalized Ornstein-Uhlenbeck process}
The Generalized Ornstein-Uhlenbeck (GOU) process is the time-varying OU process \cite{ahmad1988introduction}. It is a stationary Gaussian-Markov process, whose marginal distribution gradually tends towards a stable mean and variance over time. The GOU process is generally defined as follows:
\begin{equation}\label{eq4}
\mathrm{d}\mathbf{x}_t=\theta _t\left( \bm \mu - \mathbf{x}_t \right) \mathrm{d}t + g_t \mathrm{d}\mathbf{w}_t,
\end{equation}
where $\bm\mu$ is a given state vector, $\theta_t$ denotes a scalar drift coefficient and $g_t$ represents the diffusion coefficient. At the same time, we require $\theta_t,g_t$ to satisfy the specified relationship $2\lambda^2=g^2_t/\theta_t$, where $\lambda^2$ is a given constant scalar. As a result, its transition probability possesses a closed-form analytical solution:
\begin{equation}\label{OU_transition}
\begin{gathered}
p\left( \mathbf{x}_t \mid \mathbf{x}_s \right)
=N(\mathbf{\bar m}_{s:t},\bar \sigma_{s:t}^2\boldsymbol{I})=\\
N\left( \bm \mu +\left( \mathbf{x}_s - \bm\mu \right) e^{-\bar{\theta}_{s:t}},\frac{g^2_t}{2\theta_t}\left( 1-e^{-2\bar{\theta}_{s:t}}\right)\boldsymbol{I} \right), \\
\bar{\theta}_{s:t} = \int_s^t{\theta _zdz}.
\end{gathered}
\end{equation}
A simple proof is provided in Appendix \ref{ou}. For the sake of simplicity in subsequent representations, we denote $\bar\theta_{0:t}$ and $\bar\sigma_{0:t}$ as $\bar\theta_{t}$ and $\bar\sigma_{t}$ respectively. Consequently, $p(\mathbf{x}_t)$ will steadily converge towards a Gaussian distribution with the mean of $\bm\mu$ and the variance of $\lambda^2$ as time $t$ progresses meaning that it exhibits the mean-reverting property.

\subsection{Doob's \textit{h}-transform}
Doob's \textit{h}-transform \cite{sarkka2019applied} is a mathematical technique applied to stochastic processes. It involves transforming the original process by incorporating a specific \textit{h}-function into the drift term of the SDE, modifying the process to pass through a predetermined terminal point. More precisely, given the SDE \eqref{eq1}, if it is desired to pass through the given fixed point $\mathbf{x}_T$ at $t=T$, an additional drift term must be incorporated into the original SDE:
\begin{equation}\label{eq6}
\mathrm{d}\mathbf{x}_t=\left[\mathbf{f}( \mathbf{x}_t,t) + g^2_t\mathbf{h}(\mathbf{x}_t,t,\mathbf{x}_T,T) \right]\mathrm{d}t + g_t \mathrm{d}\mathbf{w}_t,
\end{equation}
where $\mathbf{h}(\mathbf{x}_t,t,\mathbf{x}_T,T)=\nabla_{\mathbf{x}_t}\log p(\mathbf{x}_T\mid \mathbf{x}_t)$ and $\mathbf{x}_0$ starts from $p\left( \mathbf{x}_0\mid \mathbf{x}_T \right)$. A simple proof can be found in Appendix \ref{htrans}. In comparison to \eqref{eq1}, the marginal distribution of \eqref{eq6} is conditioned on $\mathbf{x}_T$, with its forward conditional probability density given by $p(\mathbf{x}_t\mid \mathbf{x}_0, \mathbf{x}_T)$ satisfying the forward Kolmogorov equation that is defined by \eqref{eq6}. Intuitively, $p(\mathbf{x}_T \mid \mathbf{x}_0, \mathbf{x}_T) = 1$ at $t=T$, ensuring that the SDE invariably passes through the specified point $\mathbf{x}_T$ for any initial state $\mathbf{x}_0$.
\section{GOUB}
The GOU process \eqref{eq4} is characterized by mean-reverting properties that if we consider the initial state $\mathbf x_0$ to represent a high-quality image and the corresponding low-quality image $\mathbf x_T=\bm\mu$ as the final condition, then the high-quality image will gradually converge to a Gaussian distribution with the low-quality image as its mean and a stable variance $\lambda^2$.
This naturally connects some information between high and low-quality images, offering an inherent advantage in image restoration. However, the initial state of the reverse process necessitates the artificial addition of noise to low-quality images, resulting in certain information loss and thus affecting the performance \cite{luo2023image}.
 
In actuality, we are more focused on the connections between points \cite{liu2022deep,de2021diffusion,su2022dual,li2023bbdm,zhou2023denoising} in image restoration. Coincidentally, the Doob's \textit{h}-transform technique can modify an SDE such that it passes through a specified $\mathbf{x}_T$ at terminal time $T$. Accordingly, it is crucial to note that the application of the \textit{h}-transform to the GOU process effectively eliminates the impact of terminal noise, directly bridging a point-to-point relationship between high-quality and low-quality images.

\subsection{Forward and backward process}
Applying the \textit{h}-transform, we can readily derive the forward process of the GOUB, leading to the following proposition:
\begin{proposition}\label{p1}
Let $\mathbf x_t$ be a finite random variable describing by the given generalized Ornstein-Uhlenbeck process \eqref{eq4}, suppose $\mathbf{x}_T=\bm\mu$, the evolution of its marginal distribution $p(\mathbf{x}_t\mid \mathbf{x}_T)$ satisfies the following SDE:
\begin{equation}\label{eq7}
\mathrm{d}\mathbf{x}_t=\left(\theta _t + g^2_t \frac{e^{-2\bar{\theta}_{t:T}}}{\bar\sigma_{t:T}^2}  \right)(\mathbf{x}_T - \mathbf{x}_t) \mathrm{d}t + g_t \mathrm{d}\mathbf{w}_t.
\end{equation}
Additionally, the forward transition $p(\mathbf{x}_t\mid \mathbf{x}_0, \mathbf{x}_T)$ is given by:
\begin{equation}\label{forward_transition}
\begin{gathered}
p(\mathbf{x}_t\mid \mathbf{x}_0, \mathbf{x}_T)
=N(\mathbf{\bar m'}_t, \bar\sigma'^{2}_{t}\mathbf{I}),\\
\mathbf{\bar m'}_t = e^{-\bar{\theta}_{t}}\frac{\bar\sigma_{t:T}^2}{\bar\sigma_{T}^2}\mathbf{x}_0+\left[ \left(1-e^{-\bar{\theta}_{t}}\right)\frac{\bar\sigma_{t:T}^2}{\bar\sigma_{T}^2} + e^{-2\bar{\theta}_{t:T}}\frac{\bar\sigma_{t}^2}{\bar\sigma_{T}^2} \right]\mathbf{x}_T\\
\bar\sigma'^{2}_{t} = \frac{\bar\sigma_{t}^2\bar\sigma_{t:T}^2}{\bar\sigma_{T}^2}
\end{gathered}
\end{equation}
\end{proposition}
The derivation of the proposition is provided in the Appendix \ref{proofa1}. With Proposition \ref{p1}, there is no need to perform multi-step forward iteration using the SDE; instead, we can directly use its closed-form solution for one-step forward sampling.

Similarly, applying the previous SDE theory enables us to easily derive the reverse process, which leads to the following Proposition \ref{p2}:
\begin{proposition}\label{p2}
The reverse SDE of equation \eqref{eq7} has a marginal distribution $p(\mathbf{x}_t\mid \mathbf{x}_T)$, and is given by:
\begin{equation}\label{eq9}
\begin{aligned}
\mathrm{d}\mathbf{x}_t =& \Bigg[ \left(\theta _t + g^2_t \frac{e^{-2\bar{\theta}_{t:T}}}{\bar\sigma_{t:T}^2}  \right)(\mathbf{x}_T - \mathbf{x}_t) \Bigg. \\
& \Bigg. - g^2_t\nabla_{\mathbf{x}_t}\log p(\mathbf{x}_t\mid \mathbf{x}_T) \Bigg] \mathrm{d}t + g_t \mathrm{d}\mathbf{w}_t,
\end{aligned}
\end{equation}
and exists a probability flow ODE:
\begin{equation}\label{scoreode}
\begin{aligned}
\mathrm{d}\mathbf{x}_t =& \Bigg[\left(\theta _t + g^2_t \frac{e^{-2\bar{\theta}_{t:T}}}{\bar\sigma_{t:T}^2}  \right)(\mathbf{x}_T - \mathbf{x}_t)  \Bigg. \\
& - \Bigg. \frac{1}{2}g^2_t\nabla_{\mathbf{x}_t}\log p(\mathbf{x}_t\mid \mathbf{x}_T) \Bigg]\mathrm{d}t.
\end{aligned}
\end{equation}
\end{proposition}
We are capable of initiating from a low-quality image $\mathbf{x}_T$ and proceeding to utilize Euler sampling solving the reverse SDE or ODE for restoration purposes.

\subsection{Training object}
The score term $\nabla_{\mathbf{x}_t}\log p(\mathbf{x}_t\mid \mathbf{x}_T)$ can be parameterized by a neural network $\mathbf{s}_{\bm\theta}(\mathbf{x}_t,\mathbf{x}_T,t)$ and can be estimated using the loss function \eqref{eq3}. Unfortunately, training the score function for SDEs generally presents a significant challenge. Nevertheless, since the analytical form of GOUB is directly obtainable, we will introduce the use of maximum likelihood for training, which yields a more stable loss function.

We first discretize the continuous time interval $[0, T]$ into $N$ sufficiently fine-grained intervals in a reasonable manner, denoted as $\{\mathbf{x}_t\}_{t\in [0,N]},\mathbf{x}_N=\mathbf{x}_T$.
We are concerned with maximizing the log-likelihood, which leads us to the following proposition:
\begin{proposition}\label{p3}
Let $\mathbf x_t$ be a finite random variable describing by the given generalized Ornstein-Uhlenbeck process \eqref{eq4}, for a fixed $\mathbf{x}_T$, the expectation of log-likelihood $\mathbb{E}_{p(\mathbf{x}_0)} [\log p_{\bm\theta}(\mathbf{x}_{0}\mid \mathbf{x}_T)]$ possesses an Evidence Lower Bound (ELBO):
\begin{equation}\label{elbo}
\small
\begin{aligned}
&ELBO=\mathbb{E}_{p(\mathbf{x}_0)} \Bigg[ \mathbb{E} _{p\left( \mathbf{x}_1\mid \mathbf{x}_0 \right)}\left[ \log p_{\bm\theta}\left( \mathbf{x}_0\mid \mathbf{x}_1,\mathbf{x}_T \right) \right] - \Bigg. \\
& \Bigg. \sum_{t=2}^T \mathbb{E}_{p(x_t\mid x_0)}[{KL\left( p\left( \mathbf{x}_{t-1}\mid \mathbf{x}_0, \mathbf{x}_t, \mathbf{x}_T \right) ||p_{\bm\theta}\left( \mathbf{x}_{t-1}\mid \mathbf{x}_t,\mathbf{x}_T \right) \right)}]\Bigg]
\end{aligned}
\end{equation}

Assuming $ p_{\bm\theta}\left( \mathbf{x}_{t-1}\mid 
\mathbf{x}_t,\mathbf{x}_T \right)$ is a Gaussian distribution with a constant variance $N(\bm{\mu}_{\bm\theta,t-1},\sigma_{\bm\theta,t-1}^{2}\boldsymbol I)$, maximizing the ELBO is equivalent to minimizing:
\begin{equation}\label{l}
\mathcal{L}= \mathbb{E}_{t,\mathbf{x}_0,\mathbf{x}_t,\mathbf{x}_T}\left[\frac{1}{2\sigma_{\bm\theta,t-1}^{2}}\|\bm{\mu}_{t-1}-\bm{\mu}_{\bm\theta,t-1}\|^2\right],
\end{equation}
where $\bm\mu_{t-1}$ represents the mean of $p\left( \mathbf x_{t-1}\mid \mathbf x_0, \mathbf x_t, \mathbf x_T \right)$:
\begin{equation}\label{optim_mu}
\small
\bm\mu_{t-1}=\frac{1}{\bar\sigma'^{2}_{t}}\left[\bar\sigma'^{2}_{t-1}(\mathbf x_t- b \mathbf x_T)a +(\bar\sigma'^{2}_{t}-\bar\sigma'^{2}_{t-1}a^2) \mathbf{\bar m'}_t \right],
\end{equation}
where,
\begin{align*}
a&=\frac{e^{-\bar{\theta}_{t-1:t}}\bar\sigma_{t:T}^2}{\bar\sigma_{t-1:T}^2},\\
b&=\frac{1}{\bar\sigma_{T}^2}\left\{(1-e^{-\bar{\theta}_{t}})\bar\sigma^{2}_{t:T} +e^{-2\bar{\theta}_{t:T}}\bar\sigma_{t}^2 \right.\\
& \left. - \left[(1 - e^{- \bar{\theta}_{t-1}})\bar\sigma^{2}_{t-1:T} +e^{-2\bar{\theta}_{t-1:T}}\bar\sigma_{t-1}^2\right]a \right\}
\end{align*}
\end{proposition}
The derivation of the proposition is provided in the Appendix \ref{proofa3}. With Proposition \ref{p3}, we can easily construct the training objective. In this work, we try to parameterized $\bm\mu_{\bm\theta,t-1}$ from differential of SDE which can be derived from equation \eqref{eq9}:
\begin{equation}
\begin{aligned}
\mathbf x_{t-1}=&\mathbf x_{t}-\left(\theta _t + g^2_t \frac{e^{-2\bar{\theta}_{t:T}}}{\bar\sigma_{t:T}^2}  \right) (\mathbf x_T -\mathbf x_t) \\
&+ g^2_t\nabla_{\mathbf x_t}\log p(\mathbf x_t\mid \mathbf x_T)  - g_t \bm\epsilon_t,
\end{aligned}
\end{equation}
where $\bm\epsilon_t \sim N(\mathbf{0}, \mathrm{d}t \boldsymbol{I})$, therefore:
\begin{equation}
\begin{aligned}
\bm\mu_{\bm\theta,t-1} =& \mathbf x_{t}-\left(\theta _t + g^2_t \frac{e^{-2\bar{\theta}_{t:T}}}{\bar\sigma_{t:T}^2}  \right) (\mathbf x_T - \mathbf x_t) \\
&+ g^2_t\nabla_{\mathbf x_t}\log p_{\theta}(\mathbf x_t\mid \mathbf x_T),\\
\sigma_{\bm\theta,t-1}=&g_t.
\end{aligned}
\end{equation}
Inspired by conditional score matching, we can parameterize noise as $\bm\epsilon_{\bm\theta}(\mathbf x_t,\mathbf x_T,t)$, thus the score $\nabla_{\mathbf x_t}\log p_\theta(\mathbf x_t\mid \mathbf x_T)$ can be represented as $-\bm\epsilon_{\bm\theta}(\mathbf x_t, \mathbf x_T,t)/\bar\sigma'_{t}$. In addition, during our empirical research, we found that utilizing L1 loss yields enhanced image reconstruction outcomes \cite{boyd2004convex,hastie2009elements}. This approach enables the model to learn pixel-level details more easily, resulting in markedly improved visual quality. Therefore, the final training object is:
\begin{equation}
\begin{aligned}
\mathcal{L} = \mathbb{E}_{t,\mathbf x_0,\mathbf x_t,\mathbf x_T}
&\left[
\frac{1}{2g_t^2} 
\Bigg\| 
\frac{1}{\bar\sigma'^{2}_{t}}\left[\bar\sigma'^{2}_{t-1}(\mathbf x_t- b \mathbf x_T)a 
\right.\Bigg.\right.\\
&\left.\Bigg.\left. +(\bar\sigma'^{2}_{t}-\bar\sigma'^{2}_{t-1}a^2)\mathbf{\bar m'}_t \right] - \mathbf x_{t}  \right.\Bigg. \\&
\left.\Bigg. + \left(\theta _t + g^2_t \frac{e^{-2\bar{\theta}_{t:T}}}{\bar\sigma_{t:T}^2}  \right) (\mathbf x_T - \mathbf x_t) \right.\Bigg. \\
& \left.\Bigg.+ \frac{g^2_t}{\bar\sigma'_{t}}\bm\epsilon_{\bm\theta}(\mathbf x_t,\mathbf x_T,t)
\Bigg\|
\right]
\end{aligned}
\end{equation}
Consequently, if we obtain the optimal $\bm\epsilon_{\bm\theta}^{*}(\mathbf x_t, \mathbf x_T,t)$, we can compute the score $\nabla_{\mathbf x_t}\log p(\mathbf x_t\mid \mathbf x_T)\approx-\bm\epsilon_{\bm\theta}^{*}(\mathbf x_t, \mathbf x_T,t)/\bar\sigma'_{t}$ for reverse process. Starting from a low-quality image $\mathbf x_T$, we can recover $\mathbf x_0$ by using Equation \eqref{eq9} to perform reverse iteration.

\subsection{Mean-ODE}
Unlike normal diffusion models, our parameterization of the mean $\bm\mu_{\bm\theta,t-1}$ is derived from the differential of SDE which effectively combines the characteristics of discrete diffusion models and continuous score-based generative models.
In the reverse process, the value of each sampling step will approximated to the true mean during training.
Therefore, we propose a Mean-ODE model, which omits the Brownian drift term:
\begin{equation}
\begin{aligned}
\mathrm{d}\mathbf{x}_t =& \Bigg[\left(\theta _t + g^2_t \frac{e^{-2\bar{\theta}_{t:T}}}{\bar\sigma_{t:T}^2}  \right)(\mathbf{x}_T - \mathbf{x}_t)  \Bigg. \\
& - \Bigg. g^2_t\nabla_{\mathbf{x}_t}\log p(\mathbf{x}_t\mid \mathbf{x}_T) \Bigg]\mathrm{d}t,
\end{aligned}
\end{equation}
To simplify the expression, we use GOUB to represent the GOUB (SDE) sampling model and Mean-ODE to represent the GOUB (Mean-ODE) sampling model. Our following experiments have demonstrated that the Mean-ODE is more effective than the corresponding Score-ODE at capturing the pixel details and structural perceptions of images, playing a pivotal role in image restoration tasks. Concurrently, the SDE model \eqref{eq9} is more focused on deep visual features and diversity.

\section{Experiments}
 We conduct experiments under three popular image restoration tasks: image inpainting, image deraining, and image super-resolution. 
Four metrics are employed for the model evaluation, \emph{i.e.}, Peak Signal-to-Noise Ratio (PSNR) for assessing reconstruction quality, Structural Similarity Index (SSIM) \cite{wang2004image} for gauging structural perception, Learned Perceptual Image Patch Similarity (LPIPS) \cite{zhang2018unreasonable} for evaluating the depth and quality of features, and Fréchet Inception Distance (FID) \cite{heusel2017gans} to measure the diversity in generated images. 
More experiment details are present in Appendix \ref{Experimental Details}.

\begin{figure}[t]
  \vskip 0.1in
  \centering
  \includegraphics[width=0.48\textwidth]{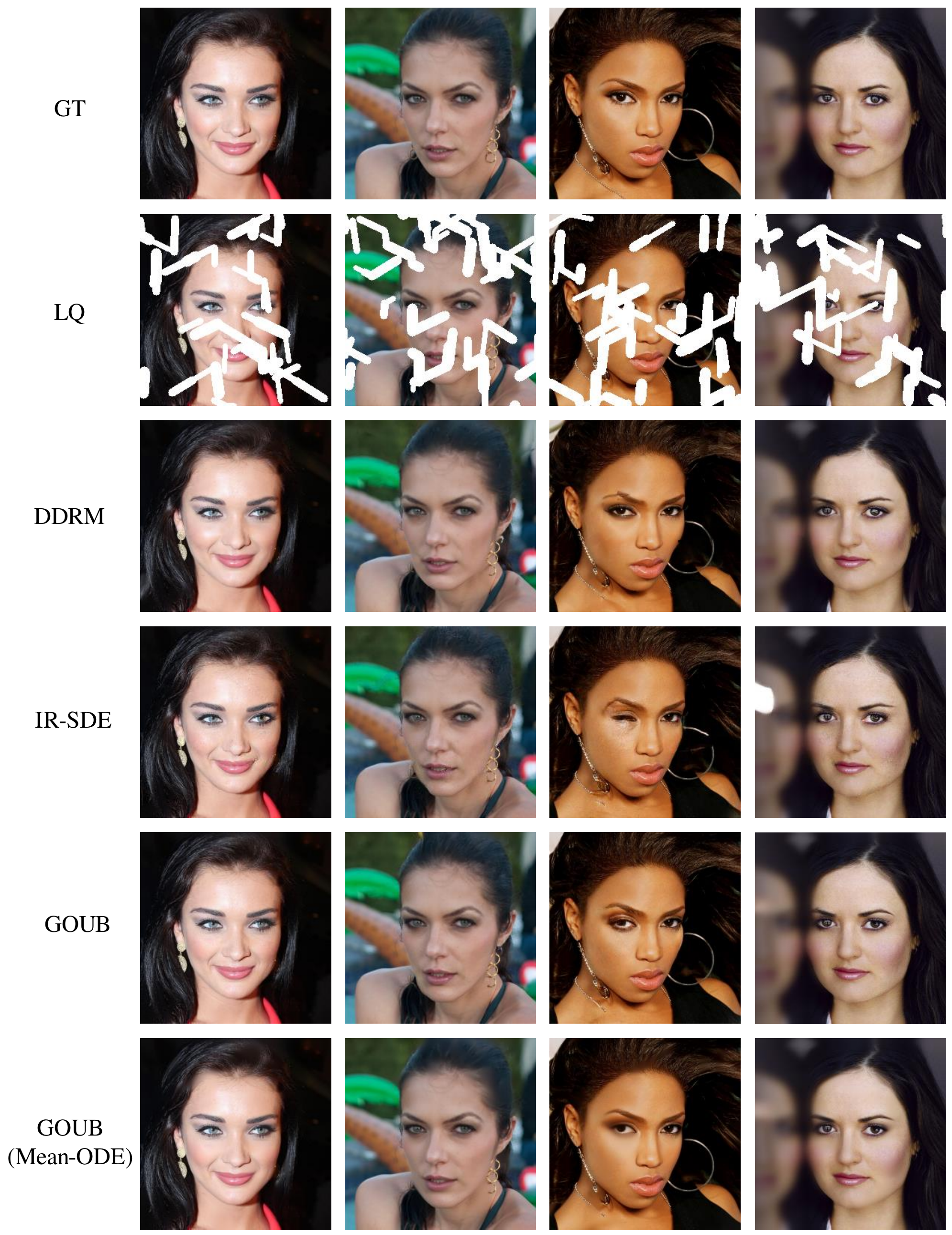}
  \caption{Qualitative comparison of the visual results of different inpainting methods on the CelebA-HQ dataset with thin mask.}
  \label{figinpaint}
  \vskip -0.2in
\end{figure}

\paragraph{Image Inpainting.} Image inpainting involves filling in missing or damaged parts of an image, to restore or enhance the overall visual effect of the image. We have selected the CelebA-HQ $256 \times 256$ datasets \cite{karras2017progressive} for both training and testing with 100 thin masks. We compare our models with several current baseline inpainting approaches such as PromptIR \cite{potlapalli2023promptir}, DDRM \cite{kawar2022denoising} and IR-SDE \cite{luo2023image}. The relevant experimental results are shown in the Table \ref{tbinpaint} and Figure \ref{figinpaint}. It is observed that the two proposed models achieved state-of-the-art results in their respective areas of strength and also delivered highly competitive outcomes on other metrics. From a visual perspective, our model excels in capturing details such as eyebrows, eyes, and image backgrounds.

\begin{table}[t]
  \centering
  \caption{\textbf{Image Inpainting.} Qualitative comparison with the relevant baselines on CelebA-HQ.}
  \vskip 0.15in
  \label{tbinpaint}
  \begin{tabular}{lcccc}
    \toprule
    \textbf{METHOD} & \textbf{PSNR\(\uparrow\)} & \textbf{SSIM\(\uparrow\)} & \textbf{LPIPS\(\downarrow\)} & \textbf{FID\(\downarrow\)} \\
    \midrule
    PromptIR & 30.22 & 0.9180 & 0.068 & 32.69 \\
    DDRM & 27.16 & 0.8993 & 0.089 & 37.02\\
    IR-SDE & 28.37 & 0.9166 & 0.046 & 25.13 \\
    \midrule
    GOUB & 28.98 & 0.9067 & \textbf{0.037} & \textbf{4.30} \\
    Mean-ODE & \textbf{31.39} & \textbf{0.9392} & 0.052 & 12.24 \\
    \bottomrule
  \end{tabular}
  \vskip -0.1in
\end{table}

\begin{figure}[ht]
\vskip 0.1in
  \centering
  \includegraphics[width=0.48\textwidth]{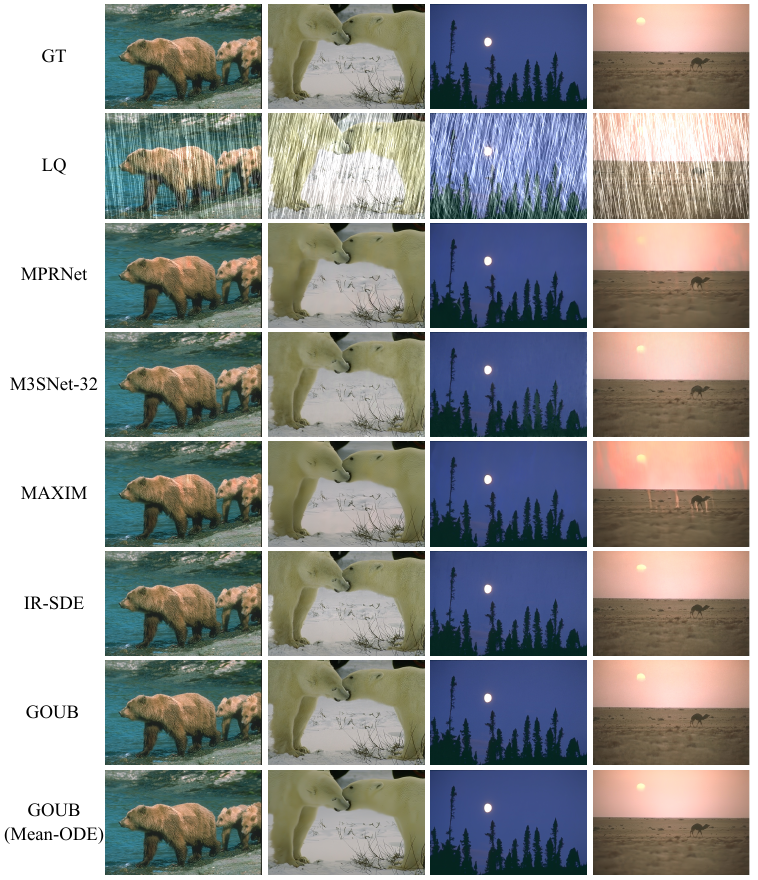}
  \caption{Qualitative comparison of the visual results of different deraining methods on the Rain100H dataset.}
  \label{figderain}
  \vskip -0.2in
\end{figure}

\paragraph{Image Deraining.}We have selected the Rain100H datasets \cite{yang2017deep} for our training and testing, which includes 1800 pairs of training data and 100 images for testing. It is important to note that in this task, similar to other deraining models, we present the PSNR and SSIM scores specifically on the Y channel (YCbCr space). We report state-of-the-art approaches for comparison: MPRNet \cite{zamir2021multi}, M3SNet-32 \cite{gao2023mountain}, MAXIM \cite{tu2022maxim}, MHNet \cite{gao2023mixed}, IR-SDE \cite{luo2023image}. The relevant experimental results are shown in the Table \ref{tbderain} and Figure \ref{figderain}. Similarly, both models achieved SOTA results respectively in the deraining task. Visually, it can be also observed that our model excels in capturing details such as the moon, the sun, and tree branches.

\begin{table}[t]
  \centering
  \caption{\textbf{Image Deraining.}  Qualitative comparison with the relevant baselines on Rain100H.}
  \label{tbderain}
  \vskip 0.15in
  \begin{tabular}{lcccc}
    \toprule
    \textbf{METHOD} & \textbf{PSNR\(\uparrow\)} & \textbf{SSIM\(\uparrow\)} & \textbf{LPIPS\(\downarrow\)} & \textbf{FID\(\downarrow\)} \\
    \midrule
    MPRNet & 30.41 & 0.8906 & 0.158 & 61.59 \\
    M3SNet-32 & 30.64 & 0.8920 & 0.154 & 60.26 \\
    MAXIM & 30.81 & 0.9027 & 0.133 & 58.72 \\
    MHNet & 31.08 & 0.8990 & 0.126 & 57.93\\
    IR-SDE & 31.65 & 0.9041 & 0.047 & 18.64 \\
    \midrule
    GOUB & 31.96 & 0.9028 & \textbf{0.046} & \textbf{18.14} \\
    Mean-ODE & \textbf{34.56} & \textbf{0.9414} & 0.077 & 32.83 \\
    \bottomrule
  \end{tabular}
  \vskip -0.1in
\end{table}

\begin{figure}[ht]
\vskip 0.1in
  \centering
  \includegraphics[width=0.48\textwidth]{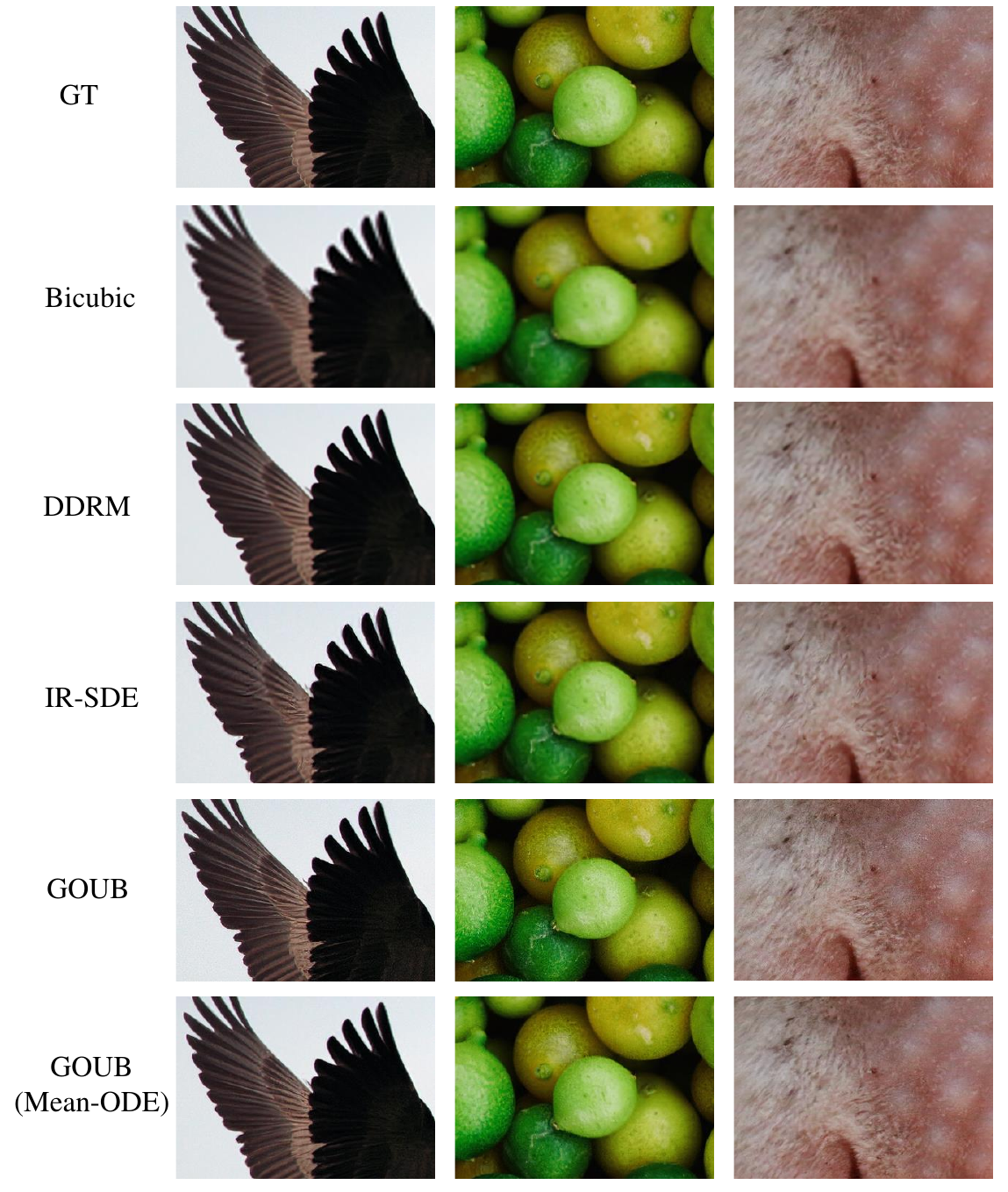}
  \caption{Qualitative comparison of the visual results of different 4x super-resolution methods on the DIV2K dataset.}
  \label{figsr}
  \vskip -0.2in
\end{figure}

\paragraph{Image Super-Resolution.}Single image super-resolution aims to recover a higher resolution and clearer version from a low-resolution image. We conducted training and evaluation on the DIV2K validation set for 4$\times$ upscaling \cite{agustsson2017ntire} and all low-resolution images were bicubically rescaled to the same size as their corresponding high-resolution images. To show that our models are in line with the state-of-the-art, we compare to the DDRM \cite{kawar2022denoising} and IR-SDE \cite{luo2023image}. The relevant experimental results are provided in Table \ref{tbsr} and Figure \ref{figsr}. As can be seen, our GOUB is superior to benchmarks in various indicators and handles visual details better such as edges and hair.

\begin{table}[t]
  \centering
  \caption{\textbf{Image 4$\times$ Super-Resolution.} Qualitative comparison with the relevant baselines on DIV2K.}
  \label{tbsr}
  \vskip 0.15in
  \begin{tabular}{lcccc}
    \toprule
    \textbf{METHOD} & \textbf{PSNR\(\uparrow\)} & \textbf{SSIM\(\uparrow\)} & \textbf{LPIPS\(\downarrow\)} & \textbf{FID\(\downarrow\)} \\
    \midrule
    DDRM & 24.35 & 0.5927 & 0.364 & 78.71\\
    IR-SDE & 25.90 & 0.6570 & 0.231 & 45.36 \\
    \midrule
    GOUB & 26.89 & 0.7478 & \textbf{0.220} & \textbf{20.85} \\
    Mean-ODE & \textbf{28.50} & \textbf{0.8070} & 0.328 & 22.14 \\
    \bottomrule
  \end{tabular}
\vskip -0.1in
\end{table}

\paragraph{Superiority of Mean-ODE.} Additionally, we conduct ablation experiments using the corresponding Score-ODE \eqref{scoreode} model to demonstrate the superiority of our proposed Mean-ODE model in image restoration. From Table \ref{comscoreode}, it is evident that the performance of Mean-ODE is significantly superior to that of the corresponding Score-ODE. This is because the sampling results of each sampling step of Mean-ODE directly approximate the true mean during the training process, as opposed to the parameterized approach such as DDPM, which relies on expectations. Consequently, our proposed Mean-ODE demonstrates better reconstruction effects and is more suitable for image restoration tasks.

\begin{table*}[t]
  \centering
  \caption{Qualitative comparison with the corresponding Score-ODE on various tasks.}
  \vskip 0.15in
  \label{comscoreode}
  \resizebox{\textwidth}{!}{
  \begin{tabular}{ccccccccccccc}
    \toprule
    \multirow{2}*{\textbf{METHOD}} & \multicolumn{4}{c}{\textbf{Image Inapinting}} & \multicolumn{4}{c}{\textbf{Image Deraining}} & \multicolumn{4}{c}{\textbf{Image 4$\times$ Super-Resolution}} \\
    \cmidrule(r){2-5} \cmidrule(r){6-9} \cmidrule(r){10-13}
      & \textbf{PSNR}$\uparrow$ & \textbf{SSIM}$\uparrow$ & \textbf{LPIPS}$\downarrow$ & \textbf{FID}$\downarrow$ & \textbf{PSNR}$\uparrow$ & \textbf{SSIM}$\uparrow$ & \textbf{LPIPS}$\downarrow$ & \textbf{FID}$\downarrow$ & \textbf{PSNR}$\uparrow$ & \textbf{SSIM}$\uparrow$ & \textbf{LPIPS}$\downarrow$ & \textbf{FID}$\downarrow$ \\
    \midrule
    Score-ODE  & 18.23 & 0.6266 & 0.389 & 161.54 & 13.64 & 0.7404 & 0.338 & 191.15 & 28.14 & 0.7993 & 0.344 & 25.51 \\
    Mean-ODE  & \textbf{31.39} & \textbf{0.9392} & \textbf{0.052} & \textbf{12.24} & 
    \textbf{34.56} & \textbf{0.9414} & \textbf{0.077} & \textbf{32.83}& 
    \textbf{28.50} & \textbf{0.8070} & \textbf{0.328} & \textbf{22.14} \\
    \bottomrule
  \end{tabular}
  }
  \vskip -0.1in
\end{table*}

\section{Analysis}
\begin{table*}[t]
  \centering
  \caption{Qualitative comparison with the different bridge models on CelebA-HQ, Rain100H, and DIV2K datasets.}
  \vskip 0.15in
  \label{tbbridge}
  \resizebox{\textwidth}{!}{
  \begin{tabular}{ccccccccccccc}
    \toprule
    \multirow{2}*{\textbf{METHOD}} & \multicolumn{4}{c}{\textbf{Image Inapinting}} & \multicolumn{4}{c}{\textbf{Image Deraining}} & \multicolumn{4}{c}{\textbf{Image 4$\times$ Super-Resolution}} \\
    \cmidrule(r){2-5} \cmidrule(r){6-9} \cmidrule(r){10-13}
      & \textbf{PSNR}$\uparrow$ & \textbf{SSIM}$\uparrow$ & \textbf{LPIPS}$\downarrow$ & \textbf{FID}$\downarrow$ & \textbf{PSNR}$\uparrow$ & \textbf{SSIM}$\uparrow$ & \textbf{LPIPS}$\downarrow$ & \textbf{FID}$\downarrow$ & \textbf{PSNR}$\uparrow$ & \textbf{SSIM}$\uparrow$ & \textbf{LPIPS}$\downarrow$ & \textbf{FID}$\downarrow$ \\
    \midrule
    VEB & 27.75 & 0.8943 & 0.056 & 13.70 & 30.39& 0.8975& 0.059 & 28.54 & 24.21&  0.5808& 0.384& 36.55 \\
    VPB  & 27.32 & 0.8841 & 0.049 & 11.87 & 30.89 & 0.8847 & 0.051 & 23.36 & 25.40 & 0.6041 & 0.342 & 29.17 \\
    GOUB & \textbf{28.98} & \textbf{0.9067} & \textbf{0.037} & \textbf{4.30} & \textbf{31.96} & \textbf{0.9028} & \textbf{0.046} & \textbf{18.14} & \textbf{26.89} & \textbf{0.7478} & \textbf{0.220} & \textbf{20.85} \\
    \bottomrule
  \end{tabular}
  }
  \vskip -0.1in
\end{table*}

\begin{figure}[t]
  \includegraphics[width=0.48\textwidth]{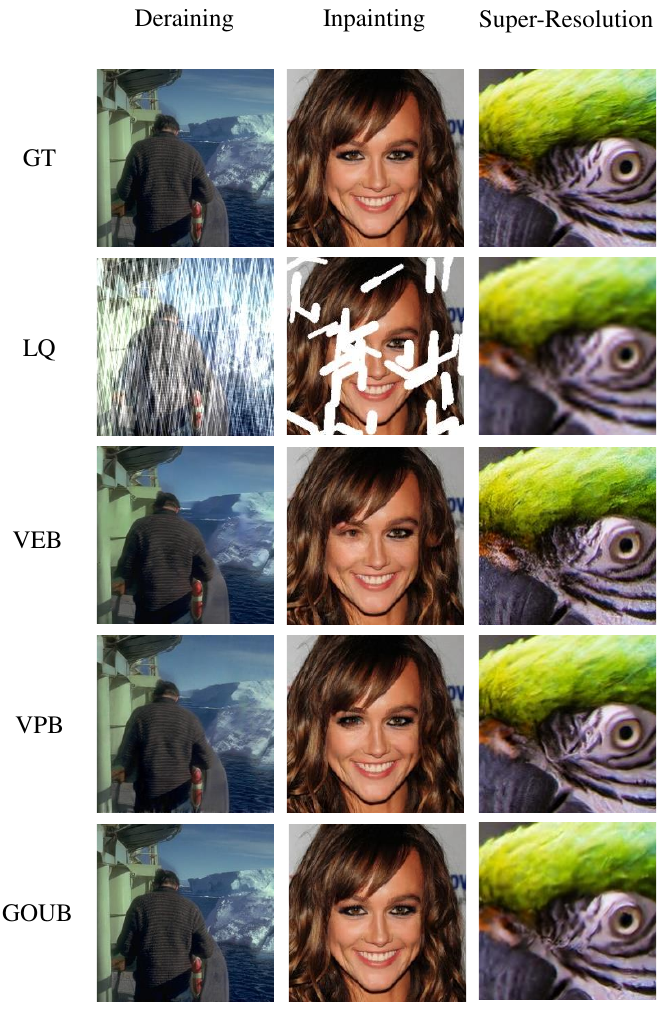}
  \caption{Qualitative comparison with the different bridge models in many tasks.}
  \vskip -0.2in
  \label{figvevpgou}
\end{figure}

The Doob's \textit{h}-transform of the generalized Ornstein-Uhlenbeck process, also known as the conditional GOU process has been an intriguing topic in previous applied mathematical research \cite{3f16a282-6308-31ec-8d78-222cd1da0f78,cheridito2003fractional,heng2021simulating}. On account of the mean-reverting property of the GOU process, applying the \textit{h}-transform makes it most straightforward to eliminate the variance and drive it towards a Dirac distribution in its steady state which is highly advantageous for its applications in image restoration. In previous research on diffusion models, there has been limited focus on the cases of $\mathbf f$ or $g$, and generally used the VE process \cite{song2020score} represented by NCSN \cite{song2019generative} or the VP process \cite{song2020score} represented by DDPM \cite{ho2020denoising}.

In this section, we demonstrate that the mathematical essence of several recent meaningful diffusion bridge models is the same \cite{li2023bbdm,zhou2023denoising,liu20232} and they all represent Brownian bridge \cite{chow2009brownian} models, details are provided in the Appendix \ref{proof_bm}. Then, we also found that the VE and VP processes are special cases of GOU, leading to the following proposition:
\begin{proposition}\label{p5}
For a given GOU process \eqref{eq4}, there exists relationships:
\begin{equation}\label{relationvevpgou}
\small
\begin{aligned}
\lim_{\theta_t \rightarrow 0} \text{GOU} = \text{VE}\\
\lim_{\bm \mu \rightarrow 0, \lambda \rightarrow 1} \text{GOU} = \text{VP}
\end{aligned}
\end{equation}
\end{proposition}
Details are provided in the Appendix \ref{vevpgou}. Therefore, we conduct experiments on VE Bridge (VEB) \cite{li2023bbdm,zhou2023denoising,liu20232} and VP Bridge (VPB) \cite{zhou2023denoising} to demonstrate the optimality of our proposed GOUB model in image restoration. We keep all the model hyperparameters consistent and results are shown in Table \ref{tbbridge} and Figure \ref{figvevpgou}.

It can be seen that under the same configuration of model hyperparameters, the performance of the GOUB is notably superior to the other two types of bridge models, which demonstrates the optimality of GOUB and also highlights the importance of the choice of diffusion process in diffusion models.

\section{Related Works}
\paragraph{Conditional Generation.} As previously highlighted, in the work of image restoration using diffusion models, the focus of some research has predominantly been on using low-quality images as conditional inputs $y$ to guide the generation process.
They \cite{kawar2021snips,saharia2022image,kawar2022denoising,chung2022diffusion,chung2022improving,chung2023parallel,zhao2023ddfm,murata2023gibbsddrm,feng2023score} all endeavor to solve or approximate the classifier $\log \nabla_{\mathbf x_t} p(\mathbf y \mid \mathbf x_t)$, necessitating the incorporation of additional prior knowledge to model specific degradation processes which both complex and lacking in universality. 

\paragraph{Diffusion Bridge.} This segment of work obviates the need for prior knowledge, constructing a diffusion bridge model from high-quality to low-quality images, thereby learning the degradation process. The previously mentioned approach \cite{liu2022deep,de2021diffusion,su2022dual,liu20232,shi2023diffusion,li2023bbdm,zhou2023denoising, albergo2023stochastic} fall into this class and are characterized by the issues of significant computational expense in solution seeking and also not the optimal model framework. Additionally, some models of flow category \cite{lipman2022flow,liu2022flow, tong2023simulation, albergo2023building, delbracio2023inversion} also belong to the diffusion bridge models and face the similar issue.

\section{Conclusion}
In this paper, we introduced the Generalized Ornstein-Uhlenbeck Bridge (GOUB) model, a diffusion bridge model that applies the Doob's \textit{h}-transform to the GOU process. This model can address general image restoration tasks without the need for specific prior knowledge. Furthermore, we have uncovered the mathematical essence of several bridge models and empirically demonstrated the optimality of our proposed model. In addition, considering our unique mean parameterization mechanism, we proposed the Mean-ODE model. Experimental results indicate that both models achieve state-of-the-art results in their respective areas of strength on various tasks, including inpainting, deraining, and super-resolution. We believe that the exploration of diffusion process and bridge models holds significant importance not only in the field of image restoration but also in advancing the study of generative diffusion models.
\section*{Impact Statements}
This paper presents work whose goal is to advance the field of Machine Learning. There are many potential societal consequences of our work, none which we feel must be specifically highlighted here.

\nocite{langley00}

\bibliography{icml2024/ref}
\bibliographystyle{icml2024/icml2024}

%%%%%%%%%%%%%%%%%%%%%%%%%%%%%%%%%%%%%%%%%%%%%%%%%%%%%%%%%%%%%%%%%%%%%%%%%%%%%%%
%%%%%%%%%%%%%%%%%%%%%%%%%%%%%%%%%%%%%%%%%%%%%%%%%%%%%%%%%%%%%%%%%%%%%%%%%%%%%%%
% APPENDIX
%%%%%%%%%%%%%%%%%%%%%%%%%%%%%%%%%%%%%%%%%%%%%%%%%%%%%%%%%%%%%%%%%%%%%%%%%%%%%%%
%%%%%%%%%%%%%%%%%%%%%%%%%%%%%%%%%%%%%%%%%%%%%%%%%%%%%%%%%%%%%%%%%%%%%%%%%%%%%%%
\newpage
\appendix
\onecolumn
\section{Proof}
\subsection{Proof of Proposition \ref{p1}}\label{proofa1}
\noindent \textbf{Proposition \ref{p1}.} 
\textit{Let $\mathbf x_t$ be a finite random variable describing by the given generalized Ornstein-Uhlenbeck process \eqref{eq4}, suppose $\mathbf{x}_T=\bm\mu$, the evolution of its marginal distribution $p(\mathbf{x}_t\mid \mathbf{x}_T)$ satisfies the following SDE:}
\begin{equation}\tag{\ref{eq7}}
\mathrm{d}\mathbf{x}_t=\left(\theta _t + g^2_t \frac{e^{-2\bar{\theta}_{t:T}}}{\bar\sigma_{t:T}^2}  \right) (\mathbf{x}_T - \mathbf{x}_t) \mathrm{d}t + g_t \mathrm{d}\mathbf{w}_t,
\end{equation}
\textit{additionally, the forward transition $p(\mathbf{x}_t\mid \mathbf{x}_0, \mathbf{x}_T)$ is given by:}
\begin{equation}\tag{\ref{forward_transition}}
\begin{aligned}
p(\mathbf{x}_t\mid \mathbf{x}_0, \mathbf{x}_T)
&=N(\mathbf{\bar m'}_t, \bar\sigma'^{2}_{t}\mathbf{I})\\
&=N\left(e^{-\bar{\theta}_{t}}\frac{\bar\sigma_{t:T}^2}{\bar\sigma_{T}^2}\mathbf{x}_0+\left[ \left(1-e^{-\bar{\theta}_{t}}\right)\frac{\bar\sigma_{t:T}^2}{\bar\sigma_{T}^2} + e^{-2\bar{\theta}_{t:T}}\frac{\bar\sigma_{t}^2}{\bar\sigma_{T}^2} \right]\mathbf{x}_T, \frac{\bar\sigma_{t}^2\bar\sigma_{t:T}^2}{\bar\sigma_{T}^2}\boldsymbol{I}\right)
\end{aligned}
\end{equation}

\textit{Proof}: Based on \eqref{OU_transition}, we have:
\begin{equation}
p\left( \mathbf{x}_t\mid \mathbf{x}_0 \right) 
=N\left( \mathbf{x}_T +\left( \mathbf{x}_0 - \mathbf{x}_T \right) e^{-\bar{\theta}_t},\bar \sigma_{t}^2\boldsymbol{I} \right)
\end{equation}

\begin{equation}
p\left( \mathbf{x}_T\mid \mathbf{x}_t \right) 
=N\left( \mathbf{x}_T +\left( \mathbf{x}_t - \mathbf{x}_T \right) e^{-\bar{\theta}_{t:T}}, \bar \sigma_{t:T}^2\boldsymbol{I} \right)
\end{equation}

\begin{equation}
p\left( \mathbf{x}_T\mid \mathbf{x}_0 \right) 
=N\left( \mathbf{x}_T +\left( \mathbf{x}_0 - \mathbf{x}_T \right) e^{-\bar{\theta}_{T}},\bar \sigma_{T}^2 \boldsymbol{I} \right)
\end{equation}

Firstly, the \textit{h} function can be directly compute:
\begin{equation}
\begin{aligned}
\mathbf{h}(\mathbf{x}_t,t,\mathbf{x}_T,T) &= \nabla_{\mathbf x_t}\log p(\mathbf x_T\mid \mathbf x_t)\\
&=-\nabla_{\mathbf x_t}\frac{\left( \mathbf{x}_t - \mathbf{x}_T \right)^2 e^{-2\bar{\theta}_{t:T}}}{2\sigma_{t:T}^2}\\
&=(\mathbf{x}_T-\mathbf{x}_t)\frac{e^{-2\bar{\theta}_{t:T}}}{\bar\sigma_{t:T}^2}
\end{aligned}
\end{equation}

Therefore, followed by Doob's \textit{h}-transform \eqref{eq6}, the SDE of marginal distribution $p(\mathbf{x}_t\mid \mathbf{x}_T)$ satisfied is :
\begin{equation}
\begin{aligned}
\mathrm{d}\mathbf{x}_t
&=\left[\mathbf{f}( \mathbf{x}_t,t) + g^2_t\mathbf{h}(\mathbf{x}_t,t,\mathbf{x}_T,T) \right]\mathrm{d}t + g_t \mathrm{d}\mathbf{w}_t\\
&=\left(\theta _t + g^2_t \frac{e^{-2\bar{\theta}_{t:T}}}{\bar\sigma_{t:T}^2}  \right)(\mathbf{x}_T - \mathbf{x}_t) \mathrm{d}t + g_t \mathrm{d}\mathbf{w}_t
\end{aligned}
\end{equation}

Furthermore, we can derive the following transition probability of $\mathbf x_t$ using Bayes' formula:
\begin{equation}
\begin{aligned}
p(\mathbf x_t \mid \mathbf x_0, \mathbf x_T) 
&=\frac{p(\mathbf x_T \mid \mathbf x_t, \mathbf x_0)p(\mathbf x_t \mid \mathbf x_0)}{p(\mathbf x_T \mid \mathbf x_0)} \\
&= \frac{p(\mathbf x_T \mid \mathbf x_t)p(\mathbf x_t \mid \mathbf x_0)}{p(\mathbf x_T \mid \mathbf x_0)}\\
\end{aligned}
\end{equation}
Since each component is independently and identically distributed (i.i.d), by considering a single dimension, we have:
\begin{equation}
\begin{aligned}
p(\mathbf x_t \mid \mathbf x_0, \mathbf x_T) 
&\propto \frac{1}{\sqrt{2\pi}\bar \sigma_{t} \bar \sigma_{t:T} / \bar \sigma_{T}}\exp{-\left\{\frac{(\mathbf x_t-[\mathbf{x}_T +\left( \mathbf{x}_0 - \mathbf{x}_T \right) e^{-\bar{\theta}_t}])^2}{2 \bar \sigma_{t}^2} + \frac{(\mathbf x_T-[\mathbf{x}_T +\left( \mathbf{x}_t - \mathbf{x}_T \right) e^{-\bar{\theta}_{t:T}}])^2}{2 \bar \sigma_{t:T}^2} \right\}}\\
&= \frac{1}{\sqrt{2\pi}\bar \sigma_{t} \bar \sigma_{t:T} / \bar \sigma_{T}}\exp{-\left\{\frac{(\mathbf x_t-[\mathbf{x}_T +\left( \mathbf{x}_0 - \mathbf{x}_T \right) e^{-\bar{\theta}_t}])^2}{2 \bar \sigma_{t}^2} + \frac{\left( \mathbf{x}_t - \mathbf{x}_T \right)^2 e^{-2\bar{\theta}_{t:T}}}{2 \bar \sigma_{t:T}^2} \right\}}\\
&\propto \frac{1}{\sqrt{2\pi}\bar \sigma_{t} \bar \sigma_{t:T} / \bar \sigma_{T}}\exp{-\left\{
\left(\frac{1}{2\bar \sigma_{t}^2} + \frac{e^{-2\bar{\theta}_{t:T}}}{2 \bar \sigma_{t:T}^2}\right)\mathbf x_t^2 - \left( \frac{\mathbf{x}_T - \left( \mathbf{x}_0 - \mathbf{x}_T \right) e^{-\bar{\theta}_t}}{\bar \sigma_{t}^2} + \frac{\mathbf{x}_Te^{-2\bar{\theta}_{t:T}}}{\bar \sigma_{t:T}^2} \right) \mathbf x_t\right\}}\\
\end{aligned}
\end{equation}

Notice that:
\begin{equation}
\begin{aligned}
\frac{1}{2\bar \sigma_{t}^2} + \frac{e^{-2\bar{\theta}_{t:T}}}{2 \bar \sigma_{t:T}^2}
&=\frac{\sigma_{t:T}^2+\bar \sigma_{t}^2 e^{-2\bar{\theta}_{t:T}}}{2\bar \sigma_{t}^2\bar \sigma_{t:T}^2} \\
&=\frac{\lambda^2 \left[(1-e^{-2\bar{\theta}_{t:T}}) + (1 -e^{-2\bar{\theta}_{t}}) e^{-2\bar{\theta}_{t:T}} \right]}{2\bar \sigma_{t}^2\bar \sigma_{t:T}^2}\\
&=\frac{\lambda^2 \left[(1-e^{-2\bar{\theta}_{t:T}}) + (e^{-2\bar{\theta}_{t:T}} -e^{-2\bar{\theta}_{T}}) \right]}{2\bar \sigma_{t}^2\bar \sigma_{t:T}^2} \\
& = \frac{\bar \sigma_{T}^2}{2\bar \sigma_{t}^2\bar \sigma_{t:T}^2}
\end{aligned}
\end{equation}
Bringing it back to (25), squaring the terms and reorganizing the equation, we obtain:

\begin{equation}
\begin{aligned}
p(\mathbf x_t \mid \mathbf x_0, \mathbf x_T) 
&\propto \frac{1}{\sqrt{2\pi}\bar \sigma_{t} \bar \sigma_{t:T} / \bar \sigma_{T}}\exp{-\left\{
\frac{\bar \sigma_{T}^2}{2\bar \sigma_{t}^2\bar \sigma_{t:T}^2} \mathbf x_t^2 - \left( \frac{\mathbf{x}_T - \left( \mathbf{x}_0 - \mathbf{x}_T \right) e^{-\bar{\theta}_t}}{\bar \sigma_{t}^2} + \frac{\mathbf{x}_Te^{-2\bar{\theta}_{t:T}}}{\bar \sigma_{t:T}^2} \right) \mathbf x_t \right\}}\\
&=\frac{1}{\sqrt{2\pi}\bar \sigma_{t} \bar \sigma_{t:T} / \bar \sigma_{T}}\exp{-\left\{\frac{\mathbf x_t^2 - \left( \left[\mathbf{x}_T - \left( \mathbf{x}_0 - \mathbf{x}_T \right) e^{-\bar{\theta}_t} \right]\frac{2\bar \sigma_{t:T}^2}{\bar \sigma_{T}^2} + e^{-2\bar{\theta}_{t:T}} \frac{2\bar \sigma_{t}^2}{\bar \sigma_{T}^2} \mathbf{x}_T \right) \mathbf x_t}{2(\bar \sigma_{t} \bar \sigma_{t:T} / \bar \sigma_{T})^2} \right\}}\\
&\propto \frac{1}{\sqrt{2\pi}\bar \sigma_{t} \bar \sigma_{t:T} / \bar \sigma_{T}}\exp-
\frac{\left\{\mathbf x_t - e^{-\bar{\theta}_{t}}\frac{\bar\sigma_{t:T}^2}{\bar\sigma_{T}^2}\mathbf{x}_0 - \left[ \left(1-e^{-\bar{\theta}_{t}}\right)\frac{\bar\sigma_{t:T}^2}{\bar\sigma_{T}^2} + e^{-2\bar{\theta}_{t:T}}\frac{\bar\sigma_{t}^2}{\bar\sigma_{T}^2} \right]\mathbf{x}_T \right\}^2}{2(\bar \sigma_{t} \bar \sigma_{t:T} / \bar \sigma_{T})^2}\\
&=N\left(e^{-\bar{\theta}_{t}}\frac{\bar\sigma_{t:T}^2}{\bar\sigma_{T}^2}\mathbf{x}_0+\left[ \left(1-e^{-\bar{\theta}_{t}}\right)\frac{\bar\sigma_{t:T}^2}{\bar\sigma_{T}^2} + e^{-2\bar{\theta}_{t:T}}\frac{\bar\sigma_{t}^2}{\bar\sigma_{T}^2} \right]\mathbf{x}_T, \frac{\bar\sigma_{t}^2\bar\sigma_{t:T}^2}{\bar\sigma_{T}^2}\boldsymbol{I}\right)
\end{aligned}
\end{equation}

This concludes the proof of the \textbf{Proposition \ref{p1}.}
\subsection{Proof of Proposition \ref{p3}}\label{proofa3}

\noindent \textbf{Proposition \ref{p3}.} 
\textit{Let $\mathbf x_t$ be a finite random variable describing by the given generalized Ornstein-Uhlenbeck process \eqref{eq4}, for a fixed $\mathbf{x}_T$, the expectation of log-likelihood $\mathbb{E}_{p(\mathbf{x}_0)} [\log p_{\bm\theta}(\mathbf{x}_{0}\mid \mathbf{x}_T)]$ possesses an Evidence Lower Bound (ELBO):}
\begin{equation}\tag{\ref{elbo}}
ELBO=\mathbb{E}_{p(\mathbf{x}_0)}\left[ \mathbb{E} _{p\left( \mathbf{x}_1\mid \mathbf{x}_0 \right)}\left[ \log p_{\bm\theta}\left( \mathbf{x}_0\mid \mathbf{x}_1,\mathbf{x}_T \right) \right] -\sum_{t=2}^T{KL\left( p\left( \mathbf{x}_{t-1}\mid \mathbf{x}_0, \mathbf{x}_t, \mathbf{x}_T \right) ||p_{\bm\theta}\left( \mathbf{x}_{t-1}\mid \mathbf{x}_t,\mathbf{x}_T \right) \right)}\right]
\end{equation}
\textit{Assuming $ p_{\bm\theta}\left( \mathbf{x}_{t-1}\mid 
\mathbf{x}_t,\mathbf{x}_T \right)$ is a Gaussian distribution with a constant variance $N(\bm{\mu}_{\bm\theta,t-1},\sigma_{\bm\theta,t-1}^{2}\boldsymbol I)$, maximizing the ELBO is equivalent to minimizing:}
\begin{equation}\tag{\ref{l}}
\mathcal{L}= \mathbb{E}_{t,\mathbf{x}_0,\mathbf{x}_t,\mathbf{x}_T}\left[\frac{1}{2\sigma_{\bm\theta,t-1}^{2}}\|\bm{\mu}_{t-1}-\bm{\mu}_{\bm\theta,t-1}\|^2\right],
\end{equation}
\textit{where $\bm\mu_{t-1}$ represents the mean of $p\left( \mathbf x_{t-1}\mid \mathbf x_0, \mathbf x_t, \mathbf x_T \right)$:}
\begin{equation}\tag{\ref{optim_mu}}
\bm\mu_{t-1}=\frac{1}{\bar\sigma'^{2}_{t}}\left[\bar\sigma'^{2}_{t-1}(\mathbf x_t- b \mathbf x_T)a +(\bar\sigma'^{2}_{t}-\bar\sigma'^{2}_{t-1}a^2) \mathbf{\bar m'}_t \right],
\end{equation}
\textit{where,}
\begin{gather*}
a=\frac{e^{-\bar{\theta}_{t-1:t}}\bar\sigma_{t:T}^2}{\bar\sigma_{t-1:T}^2},\\
b=\frac{1}{\bar\sigma_{T}^2}\left\{(1-e^{-\bar{\theta}_{t}})\bar\sigma^{2}_{t:T} +e^{-2\bar{\theta}_{t:T}}\bar\sigma_{t}^2 - \left[(1 - e^{- \bar{\theta}_{t-1}})\bar\sigma^{2}_{t-1:T} +e^{-2\bar{\theta}_{t-1:T}}\bar\sigma_{t-1}^2\right]a \right\}
\end{gather*}

\textit{Proof}: Firstly, followed by the theorem in DDPM \cite{ho2020denoising}:
\begin{equation}
\begin{aligned}
\mathbb{E}_{p(\mathbf{x}_0)}\left[\log p_{\bm\theta}(\mathbf x_0) \right] \geq & \mathbb{E}_{p(\mathbf{x}_0)}\Bigg[ -KL(p(\mathbf x_T\mid \mathbf x_0)|| p(\mathbf x_T)) + \mathbb{E} _{p\left( \mathbf{x}_1\mid \mathbf{x}_0 \right)}\left[ \log p_{\bm\theta}\left( \mathbf{x}_0\mid \mathbf{x}_1 \right) \right]  \Bigg. \\ & \Bigg. - \sum_{t=2}^T \mathbb{E}_{p(x_t\mid x_0)}[{KL\left( p\left( \mathbf{x}_{t-1}\mid \mathbf{x}_0, \mathbf{x}_t \right) ||p_{\bm\theta}\left( \mathbf{x}_{t-1}\mid \mathbf{x}_t \right) \right)}]\Bigg]
\end{aligned}
\end{equation}

Similarly, we have:
\begin{equation}
\begin{aligned}
\mathbb{E}_{p(\mathbf{x}_0)} [\log p_{\bm\theta}(\mathbf{x}_{0}\mid \mathbf{x}_T)] &\geq \mathbb{E}_{p(\mathbf{x}_0)}\Bigg[ -KL(p(\mathbf x_T\mid \mathbf x_0, \mathbf{x}_T)|| p(\mathbf x_T\mid \mathbf{x}_T)) + \mathbb{E} _{p\left( \mathbf{x}_1\mid \mathbf{x}_0 \right)}\left[ \log p_{\bm\theta}\left( \mathbf{x}_0\mid \mathbf{x}_1, \mathbf{x}_T \right) \right]  \Bigg. \\ & \Bigg. \quad - \sum_{t=2}^T \mathbb{E}_{p(x_t\mid x_0)} [{KL\left( p\left( \mathbf{x}_{t-1}\mid \mathbf{x}_0, \mathbf{x}_t, \mathbf{x}_T \right) ||p_{\bm\theta}\left( \mathbf{x}_{t-1}\mid \mathbf{x}_t, \mathbf{x}_T \right) \right)}]\Bigg]\\
&=
\mathbb{E}_{p(\mathbf{x}_0)}\Bigg[ \mathbb{E} _{p\left( \mathbf{x}_1\mid \mathbf{x}_0 \right)}\left[ \log p_{\bm\theta}\left( \mathbf{x}_0\mid \mathbf{x}_1,\mathbf{x}_T \right) \right]
\Bigg. \\ & \Bigg.
\quad -\sum_{t=2}^T \mathbb{E}_{p(x_t\mid x_0)}[{KL\left( p\left( \mathbf{x}_{t-1}\mid \mathbf{x}_0, \mathbf{x}_t, \mathbf{x}_T \right) ||p_{\bm\theta}\left( \mathbf{x}_{t-1}\mid \mathbf{x}_t,\mathbf{x}_T \right) \right)}]\Bigg]
\\
&=ELBO
\end{aligned}
\end{equation}

From Bayes' formula, we can infer that:
\begin{equation}\label{byesi}
\begin{aligned}
p\left( \mathbf{x}_{t-1}\mid \mathbf{x}_0, \mathbf{x}_t, \mathbf{x}_T \right)
&=\frac{p(\mathbf x_t \mid \mathbf x_0, \mathbf x_{t-1}, \mathbf x_T) p(\mathbf x_{t-1} \mid \mathbf x_0, \mathbf x_T)}{p(\mathbf x_t \mid \mathbf x_0, \mathbf x_T)} \\
&= \frac{p(\mathbf x_t \mid \mathbf x_{t-1}, \mathbf x_T)p(\mathbf x_{t-1} \mid \mathbf x_0, \mathbf x_T)}{p(\mathbf x_t \mid \mathbf x_0, \mathbf x_T)}\\
\end{aligned}
\end{equation}

Since $p(\mathbf x_{t-1} \mid \mathbf x_0, \mathbf x_T)$ and $p(\mathbf x_t \mid \mathbf x_0, \mathbf x_T)$ are Gaussian distributions \eqref{forward_transition}, by employing the reparameterization technique:
\begin{equation}
\begin{aligned}
\mathbf x_{t-1}&=e^{-\bar{\theta}_{t-1}}\frac{\bar\sigma_{t-1:T}^2}{\bar\sigma_{T}^2}\mathbf{x}_0+\left[ \left(1-e^{-\bar{\theta}_{t-1}}\right)\frac{\bar\sigma_{t-1:T}^2}{\bar\sigma_{T}^2} + e^{-2\bar{\theta}_{t-1:T}}\frac{\bar\sigma_{t-1}^2}{\bar\sigma_{T}^2} \right]\mathbf{x}_T+ \bar\sigma'_{t-1} \bm\epsilon_{t-1}\\
&=m(t-1) \mathbf{x}_0+ n(t-1)\mathbf{x}_T + \bar\sigma'_{t-1} \bm\epsilon_{t-1}
\\
\mathbf x_{t}&= e^{-\bar{\theta}_{t}}\frac{\bar\sigma_{t:T}^2}{\bar\sigma_{T}^2}\mathbf{x}_0+\left[ \left(1-e^{-\bar{\theta}_{t}}\right)\frac{\bar\sigma_{t:T}^2}{\bar\sigma_{T}^2} + e^{-2\bar{\theta}_{t:T}}\frac{\bar\sigma_{t}^2}{\bar\sigma_{T}^2} \right]\mathbf{x}_T + \bar\sigma'_{t} \bm\epsilon_t\\
&=m(t) \mathbf{x}_0+ n(t)\mathbf{x}_T + \bar\sigma'_{t} \bm\epsilon_t
\end{aligned}
\end{equation}

Therefore,
\begin{equation}
\begin{aligned}
\mathbf x_{t} &=  \frac{m(t)}{m(t-1)} \mathbf x_{t-1} + \left[n(t)-\frac{m(t)}{m(t-1)}n(t-1) \right]\mathbf x_{T} + \sqrt{\bar\sigma'^2_{t}-\frac{m(t)^2}{m(t-1)^2}\bar\sigma'^2_{t-1}} \bm\epsilon\\
&= a \mathbf x_{t-1} + \left[n(t)-an(t-1) \right]\mathbf x_{T} + \sqrt{\bar\sigma'^2_{t}-a^2\bar\sigma'^2_{t-1}} \bm\epsilon\\
&= a \mathbf x_{t-1} + b\mathbf x_{T} + \sqrt{\bar\sigma'^2_{t}-a^2\bar\sigma'^2_{t-1}} \bm\epsilon
\end{aligned}
\end{equation}

Thus, $p(\mathbf x_t \mid \mathbf x_{t-1}, \mathbf x_T)=N(a \mathbf x_{t-1} + b\mathbf x_{T},\left(\bar\sigma'^2_{t}-a^2\bar\sigma'^2_{t-1}\right)\boldsymbol I)$ is also a Gaussian distribution. Bring it back to equation \eqref{byesi} we can easily obtain :
\begin{equation}\tag{\ref{optim_mu}}
\bm\mu_{t-1}=\frac{1}{\bar\sigma'^{2}_{t}}\left[\bar\sigma'^{2}_{t-1}(\mathbf x_t- b \mathbf x_T)a +(\bar\sigma'^{2}_{t}-\bar\sigma'^{2}_{t-1}a^2) \mathbf{\bar m'}_t \right],
\end{equation}

Accordingly, 
\begin{equation}
\begin{aligned}
&KL\left( p\left( \mathbf{x}_{t-1}\mid \mathbf{x}_0, \mathbf{x}_t, \mathbf{x}_T \right) ||p_{\bm\theta}\left( \mathbf{x}_{t-1}\mid \mathbf{x}_t,\mathbf{x}_T \right) \right)\\
=&\mathbb{E}_{p\left( \mathbf{x}_{t-1}\mid \mathbf{x}_0, \mathbf{x}_t, \mathbf{x}_T \right)}\left[\log \frac{ \frac{1}{\sqrt{2\pi}\sigma_{t-1}}e^{-(\mathbf x_{t-1}-\bm\mu_{t-1})^2/{2\sigma_{t-1}^2}} } {\frac{1}{\sqrt{2\pi}\sigma_{\bm\theta,t-1}}e^{-(\mathbf x_{t-1}-\bm{\mu}_{\bm\theta,t-1})^2/{2\sigma_{\bm\theta,t-1}^2}}} \right]\\ 
=&\mathbb{E}_{p\left( \mathbf{x}_{t-1}\mid \mathbf{x}_0, \mathbf{x}_t, \mathbf{x}_T \right)}\left[\log\sigma_{\bm\theta,t-1} - \log\sigma_{t-1} - (\mathbf x_{t-1}-\bm\mu_{t-1})^2/{2\sigma^{2}_{t-1}} + (\mathbf x_{t-1}-\bm{\mu}_{\bm\theta,t-1})^2/{2\sigma_{\bm\theta,t-1}^{2}} \right]\\ 
=&\log\sigma_{\bm\theta,t-1}-\log\sigma_{t-1}-\frac{1}{2} + \frac{\sigma_{t-1}^2}{2\sigma_{\bm\theta,t-1}^2} + \frac{(\bm\mu_{t-1}-\bm{\mu}_{\bm\theta,t-1})^2}{2\sigma_{\bm\theta,t-1}^2}
\end{aligned}
\end{equation}
Ignoring unlearnable constant, the training object that involves minimizing the negative ELBO is :
\begin{equation}
\mathcal{L}= \mathbb{E}_{t,\mathbf{x}_0,\mathbf{x}_t,\mathbf{x}_T}\left[\frac{1}{2\sigma_{\bm\theta,t-1}^{2}}\|\bm{\mu}_{t-1}-\bm{\mu}_{\bm\theta,t-1}\|^2\right],
\end{equation}
This concludes the proof of the \textbf{Proposition \ref{p3}.}
\section{Theoretical Results}
\subsection{Brownian Bridge}
\label{proof_bm}
In this section, we will show the mathematical essence of some other bridge models, some of which are all equivalent.

\begin{proposition}\label{p4}
The mathematical essence of BBDM \cite{li2023bbdm}, DDBM (VE) \cite{zhou2023denoising} and $I^2$SB \cite{liu20232} are all equivalent to the Brownian bridge.
\end{proposition}

\textit{Proof}: Firstly, it is easy to understand that BBDM uses the Brownian bridge as its fundamental model architecture. 

The DDBM (VE) model is derived as the Doob's \textit{h}--transform of VE-SDE, and we begin by specifying the SDE:
\begin{equation}\label{35}
\mathrm{d}\mathbf{x}_t=  \mathrm{d}\mathbf{w}_t
\end{equation}
Its transition probability is given by:
\begin{equation}
p\left( \mathbf{x}_t \mid \mathbf{x}_s \right) = N(\mathbf x_s, t-s)
\end{equation}
Since, the \textit{h}--function of SDE \eqref{35} is:
\begin{equation}
\begin{aligned}
\mathbf{h}(\mathbf{x}_t,t,\mathbf{x}_T,T) 
&= \nabla_{\mathbf x_t}\log p(\mathbf x_T\mid \mathbf x_t)\\
&= \frac{\mathbf x_T-\mathbf x_t}{T-t}
\end{aligned}
\end{equation}

Therefore, the Doob's \textit{h}--transform of \eqref{35} is:
\begin{equation}\label{38}
\begin{aligned}
\mathrm{d}\mathbf{x}_t= \frac{\mathbf x_T-\mathbf x_t}{T-t}\mathrm dt + \mathrm{d}\mathbf{w}_t
\end{aligned}
\end{equation}
That is the definition of Brownian bridge. Hence, DDBM (VE) is a Brownian bridge model.

Furthermore, the transition kernel of \eqref{38} is:
\begin{equation}
\begin{aligned}
p(\mathbf x_t \mid \mathbf x_0, \mathbf x_T) 
&=\frac{p(\mathbf x_T \mid \mathbf x_t, \mathbf x_0)p(\mathbf x_t \mid \mathbf x_0)}{p(\mathbf x_T \mid \mathbf x_0)} \\
&= \frac{p(\mathbf x_T \mid \mathbf x_t)p(\mathbf x_t \mid \mathbf x_0)}{p(\mathbf x_T \mid \mathbf x_0)}\\
&= \frac{N(\mathbf x_t, T-t)N(\mathbf x_0, t)}{N(\mathbf x_0, T)}\\
&= N\left(\left(1-\frac{t}{T} \right) \mathbf x_0 + \frac{t}{T} \mathbf x_T, \frac{t(T-t)}{T}\boldsymbol{I}\right)
\end{aligned}
\end{equation}
This precisely corresponds to the sampling process of $\mathrm I^2$SB, thus confirming that $\mathrm I^2$SB also represents a Brownian bridge.

This concludes the proof of the \textbf{Proposition \ref{p4}.}

\subsection{Connections Between GOU, VE and VP}\label{vevpgou}
The following proposition will show us that both VE and VP processes are special cases of GOU process:

\noindent \textbf{Proposition \ref{p5}.}
\textit{For a given GOU process \eqref{eq4}, there exists relationships:}
\begin{equation}\tag{\ref{relationvevpgou}}
\begin{aligned}
\lim_{\theta_t \rightarrow 0} \text{GOU} = \text{VE}\\
\lim_{\bm \mu \rightarrow 0, \lambda \rightarrow 1} \text{GOU} = \text{VP}
\end{aligned}
\end{equation}

\textit{Proof}: It's easy to know:
\begin{equation}
\begin{aligned}
\lim_{\theta_t \rightarrow 0} \text{GOU} 
&= \lim_{\theta_t \rightarrow 0} \left\{\mathrm{d}\mathbf{x}_t=\theta _t\left( \bm \mu - \mathbf{x}_t \right) \mathrm{d}t + g_t \mathrm{d}\mathbf{w}_t \right\} \\
&= \lim_{\theta_t \rightarrow 0} \left\{\mathrm{d}\mathbf{x}_t= g_t \mathrm{d}\mathbf{w}_t \right\} \\
&= \text{VE},
\end{aligned}
\end{equation}
where $g_t$ will be controlled by $\lambda^2$.

Besides, we have:
\begin{equation}
\begin{aligned}
\lim_{\bm \mu \rightarrow 0, \lambda \rightarrow 1} \text{GOU} 
&= \lim_{\bm \mu \rightarrow 0, \lambda \rightarrow 1} \left\{\mathrm{d}\mathbf{x}_t=\theta _t\left( \bm \mu - \mathbf{x}_t \right) \mathrm{d}t + g_t \mathrm{d}\mathbf{w}_t \right\} \\
&= \lim_{\bm \mu \rightarrow 0, \lambda \rightarrow 1} \left\{\mathrm{d}\mathbf{x}_t=\theta _t  \bm \mu  \mathrm{d}t - \theta _t \mathbf{x}_t  \mathrm{d}t + g_t \mathrm{d}\mathbf{w}_t \right\} \\
&= \lim_{\bm \mu \rightarrow 0, \lambda \rightarrow 1} \left\{\mathrm{d}\mathbf{x}_t= - \frac{1}{2}g_t^2 \mathbf{x}_t  \mathrm{d}t + g_t \mathrm{d}\mathbf{w}_t \right\} \\
&= \text{VP},
\end{aligned}
\end{equation}
where $g_t$ will be controlled by $\theta_t$.

This concludes the proof of the \textbf{Proposition \ref{p5}.}
\section{GOU Process}\label{ou}
\begin{theorem}\label{t1}
For a given GOU process:
\begin{equation}\tag{\ref{eq4}}
\mathrm{d}\mathbf{x}_t=\theta _t\left( \bm \mu - \mathbf{x}_t \right) \mathrm{d}t + g_t \mathrm{d}\mathbf{w}_t
\end{equation}
where $\bm\mu$ is a given state vector, $\theta_t$ denotes a scalar drift coefficient and $g_t$ represents the diffusion coefficient. It possesses a closed-form analytical solution:
\begin{equation}\tag{\ref{OU_transition}}
p\left( \mathbf{x}_t \mid \mathbf{x}_s \right) 
=N\left( \bm \mu +\left( \mathbf{x}_s - \bm\mu \right) e^{-\bar{\theta}_{s:t}},\frac{g^2_t}{2\theta_t}\left( 1-e^{-2\bar{\theta}_{s:t}}\right)\boldsymbol{I} \right), \qquad \bar{\theta}_{s:t} = \int_s^t{\theta _zdz}
\end{equation}
\end{theorem}
\textit{Proof}: Writing:
\begin{equation}
\mathbf f(\mathbf x_t,t)= \mathbf x_t e^{\bar\theta_t}
\end{equation}
Using Ito differential formula, we get:
\begin{equation}
\begin{aligned}
\mathrm{d} \mathbf f(\mathbf x_t,t)
&= \mathbf x_t \theta_t e^{\bar\theta_t} \mathrm{d}t + e^{\bar\theta_t} \mathrm{d}\mathbf x_t\\
&= \mathbf x_t \theta_t e^{\bar\theta_t} \mathrm{d}t + e^{\bar\theta_t} \left[\theta _t\left( \bm \mu - \mathbf{x}_t \right) \mathrm{d}t + g_t \mathrm{d}\mathbf{w}_t \right]\\
&= e^{\bar\theta_t}\theta_t\bm \mu + e^{\bar\theta_t} g_t \mathrm{d}\mathbf{w}_t
\end{aligned}
\end{equation}
Integrating from $s$ to $t$ we get:
\begin{equation}
\begin{aligned}
\mathbf x_t e^{\bar\theta_t} - \mathbf x_s e^{\bar\theta_s} 
&= \int_{s}^{t} e^{\bar\theta_z}\theta_z\bm \mu \mathrm{d}z + \int_{s}^{t} e^{\bar\theta_z} g_z \mathrm{d}\mathbf{w}_z\\
&= \left(e^{\bar\theta_t}-e^{\bar\theta_s} \right)\bm\mu  + \int_{s}^{t} e^{\bar\theta_z} g_z \mathrm{d}\mathbf{w}_z\\
\end{aligned}
\end{equation}
It's obvious that the transition kernel is a Gaussian distribution. Since $\mathrm{d}\mathbf{w}_z\sim N(\mathbf 0,\mathrm{d}z\boldsymbol{I})$, we have:
\begin{equation}
\begin{aligned}
\int_{s}^{t} e^{\bar\theta_z} g_z \mathrm{d}\mathbf{w}_z
&=N\left(\mathbf 0,\int_{s}^{t} e^{2\bar\theta_z} g^2_z \mathrm{d}z \boldsymbol{I}\right)\\
&=N\left(\mathbf 0, \lambda^2\int_{s}^{t} e^{2\bar\theta_z} 2\theta_t \mathrm{d}z \boldsymbol{I}\right)\\
&=N\left(\mathbf 0, \lambda^2 \left(e^{2\bar\theta_t} - e^{2\bar\theta_s}\right) \boldsymbol{I}\right)\\
\end{aligned}
\end{equation}
Therefore:
\begin{equation}
\begin{aligned}
\mathbf x_t e^{\bar\theta_t} - \mathbf x_s e^{\bar\theta_s} = \left(e^{\bar\theta_t}-e^{\bar\theta_s} \right)\bm\mu  + N\left(\mathbf 0, \lambda^2 \left(e^{2\bar\theta_t} - e^{2\bar\theta_s}\right) \boldsymbol{I}\right)\\
\mathbf x_t= \bm \mu +\left( \mathbf{x}_s - \bm\mu \right) e^{-\bar{\theta}_{s:t}} + N\left(\mathbf 0, \frac{g^2_t}{2\theta_t}\left( 1-e^{-2\bar{\theta}_{s:t}}\right)\boldsymbol{I} \right)
\end{aligned}
\end{equation}
This concludes the proof of the \textbf{Theorem \ref{t1}.}
\section{Doob's \textit{h}--transform}\label{htrans}
\begin{theorem}\label{t2}
For a given SDE:
\begin{equation}\tag{\ref{eq1}}
\mathrm{d}\mathbf{x}_t=\mathbf{f}\left( \mathbf{x}_t,t \right) \mathrm{d}t+g_t \mathrm{d}\mathbf{w}_t,\qquad \mathbf{x}_0\sim p\left( \mathbf{x}_0 \right), 
\end{equation}
For a fixed $\mathbf{x}_T$, the evolution of conditional probability $p(\mathbf{x}_t\mid \mathbf{x}_T)$ follows:
\begin{equation}\tag{\ref{eq6}}
\mathrm{d}\mathbf{x}_t=\left[\mathbf{f}( \mathbf{x}_t,t) + g^2_t\mathbf{h}(\mathbf{x}_t,t,\mathbf{x}_T,T) \right]\mathrm{d}t + g_t \mathrm{d}\mathbf{w}_t,\qquad \mathbf{x}_0\sim p\left( \mathbf{x}_0 \mid\mathbf{x}_T \right),
\end{equation}
where $\mathbf{h}(\mathbf{x}_t,t,\mathbf{x}_T,T)=\nabla_{\mathbf{x}_t}\log p(\mathbf{x}_T\mid \mathbf{x}_t)$.
\end{theorem}
\textit{Proof}:
$p(\mathbf x_t \mid \mathbf x_0)$ satisfies Kolmogorov Forward Equation (KFE) also called Fokker-Planck equation \cite{risken1996fokker}:
\begin{equation}
\frac{\partial}{\partial t} p(\mathbf x_t \mid \mathbf x_0) = -\nabla_{\mathbf x_t} \cdot \left[ \mathbf f(\mathbf x_t, t)p(\mathbf x_t \mid \mathbf x_0) \right] + \frac{1}{2}g^2_t \nabla_{\mathbf x_t} \cdot \nabla_{\mathbf x_t} p(\mathbf x_t \mid \mathbf x_0)
\end{equation}
Similarly, $p(\mathbf x_T \mid \mathbf x_t)$ satisfies Kolmogorov Backward Equation (KBE) \cite{risken1996fokker}:
\begin{equation}
-\frac{\partial}{\partial t} p(\mathbf x_T \mid \mathbf x_t) = \mathbf f(\mathbf x_t, t) \cdot \nabla_{\mathbf x_t} p(\mathbf x_T \mid \mathbf x_t) + \frac{1}{2}g^2_t \nabla_{\mathbf x_t} \cdot \nabla_{\mathbf x_t} p(\mathbf x_T\mid \mathbf x_t)
\end{equation}
Using Bayes' rule, we have:
\begin{equation}
\begin{aligned}
p(\mathbf x_t \mid \mathbf x_0, \mathbf x_T) 
&=\frac{p(\mathbf x_T \mid \mathbf x_t, \mathbf x_0)p(\mathbf x_t \mid \mathbf x_0)}{p(\mathbf x_T \mid \mathbf x_0)} \\
&= \frac{p(\mathbf x_T \mid \mathbf x_t)p(\mathbf x_t \mid \mathbf x_0)}{p(\mathbf x_T \mid \mathbf x_0)}\\
\end{aligned}
\end{equation}
Therefore, the derivative of conditional transition probability $p(\mathbf x_t \mid \mathbf x_0, \mathbf x_T)$ with time follows:
\begin{equation}\label{51}
\begin{aligned}
\frac{\partial}{\partial t} p(\mathbf x_t \mid \mathbf x_0, \mathbf x_T) 
&= \frac{p(\mathbf x_t \mid \mathbf x_0)}{p(\mathbf x_T \mid \mathbf x_0)}\frac{\partial}{\partial t} p(\mathbf x_T \mid \mathbf x_t) + \frac{p(\mathbf x_T \mid \mathbf x_t)}{p(\mathbf x_T \mid \mathbf x_0)} \frac{\partial}{\partial t} p(\mathbf x_t \mid \mathbf x_0)\\
&= \frac{p(\mathbf x_t \mid \mathbf x_0)}{p(\mathbf x_T \mid \mathbf x_0)}\left[-\mathbf f(\mathbf x_t, t) \cdot \nabla_{\mathbf x_t} p(\mathbf x_T \mid \mathbf x_t) - \frac{1}{2}g^2_t \nabla_{\mathbf x_t} \cdot \nabla_{\mathbf x_t} p(\mathbf x_T\mid \mathbf x_t) \right] 
\\&\quad + \frac{p(\mathbf x_T \mid \mathbf x_t)}{p(\mathbf x_T \mid \mathbf x_0)} \left\{ -\nabla_{\mathbf x_t} \cdot \left[ \mathbf f(\mathbf x_t, t)p(\mathbf x_t \mid \mathbf x_0) \right] + \frac{1}{2}g^2_t \nabla_{\mathbf x_t} \cdot \nabla_{\mathbf x_t} p(\mathbf x_t \mid \mathbf x_0) \right\}\\
&= - \left[\frac{p(\mathbf x_t \mid \mathbf x_0)}{p(\mathbf x_T \mid \mathbf x_0)}\mathbf f(\mathbf x_t, t) \cdot \nabla_{\mathbf x_t} p(\mathbf x_T \mid \mathbf x_t) + \frac{p(\mathbf x_T \mid \mathbf x_t)}{p(\mathbf x_T \mid \mathbf x_0)} \mathbf f(\mathbf x_t, t)\nabla_{\mathbf x_t} p(\mathbf x_t\mid \mathbf x_0) \right.\\
&\quad \left. + \frac{p(\mathbf x_T \mid \mathbf x_t)}{p(\mathbf x_T \mid \mathbf x_0)}p(\mathbf x_t\mid \mathbf x_0) \nabla_{\mathbf x_t} \cdot \mathbf f(\mathbf x_t, t)\right]\\
&\quad+ \frac{1}{2}g_t^2 \left[\frac{p(\mathbf x_T \mid \mathbf x_t)}{p(\mathbf x_T \mid \mathbf x_0)} \nabla_{\mathbf x_t} \cdot \nabla_{\mathbf x_t} p(\mathbf x_t \mid \mathbf x_0)- \frac{p(\mathbf x_t \mid \mathbf x_0)}{p(\mathbf x_T \mid \mathbf x_0)}\nabla_{\mathbf x_t} \cdot \nabla_{\mathbf x_t} p(\mathbf x_T\mid \mathbf x_t) \right]\\
&=-\left[\mathbf f(\mathbf x_t, t) \cdot \nabla_{\mathbf x_t} p(\mathbf x_t \mid \mathbf x_0, \mathbf x_T) + p(\mathbf x_t \mid \mathbf x_0, \mathbf x_T)\cdot \nabla_{\mathbf x_t} \mathbf f(\mathbf x_t, t) \right]\\
&\quad+\frac{1}{2}g_t^2 \left[\frac{p(\mathbf x_T \mid \mathbf x_t)}{p(\mathbf x_T \mid \mathbf x_0)} \nabla_{\mathbf x_t} \cdot \nabla_{\mathbf x_t} p(\mathbf x_t \mid \mathbf x_0)- \frac{p(\mathbf x_t \mid \mathbf x_0)}{p(\mathbf x_T \mid \mathbf x_0)}\nabla_{\mathbf x_t} \cdot \nabla_{\mathbf x_t} p(\mathbf x_T\mid \mathbf x_t) \right]\\
&= - \nabla_{\mathbf x_t} \cdot \left[\mathbf f(\mathbf x_t, t) p(\mathbf x_t \mid \mathbf x_0, \mathbf x_T) \right]\\
&\quad+\frac{1}{2}g_t^2 \left[\frac{p(\mathbf x_T \mid \mathbf x_t)}{p(\mathbf x_T \mid \mathbf x_0)} \nabla_{\mathbf x_t} \cdot \nabla_{\mathbf x_t} p(\mathbf x_t \mid \mathbf x_0)- \frac{p(\mathbf x_t \mid \mathbf x_0)}{p(\mathbf x_T \mid \mathbf x_0)}\nabla_{\mathbf x_t} \cdot \nabla_{\mathbf x_t} p(\mathbf x_T\mid \mathbf x_t) \right]
\end{aligned}
\end{equation}
For the second term, we have:
\begin{equation}
\begin{aligned}
&\frac{1}{2}g_t^2 \left[\frac{p(\mathbf x_T \mid \mathbf x_t)}{p(\mathbf x_T \mid \mathbf x_0)} \nabla_{\mathbf x_t} \cdot \nabla_{\mathbf x_t} p(\mathbf x_t \mid \mathbf x_0)
- \frac{p(\mathbf x_t \mid \mathbf x_0)}{p(\mathbf x_T \mid \mathbf x_0)}\nabla_{\mathbf x_t} \cdot \nabla_{\mathbf x_t} p(\mathbf x_T\mid \mathbf x_t) \right]\\
=&\frac{1}{2}g_t^2\left[\frac{p(\mathbf x_T \mid \mathbf x_t)}{p(\mathbf x_T \mid \mathbf x_0)} \nabla_{\mathbf x_t} \cdot \nabla_{\mathbf x_t} p(\mathbf x_t \mid \mathbf x_0)+\frac{1}{p(\mathbf x_T \mid \mathbf x_0)}\nabla_{\mathbf x_t} p(\mathbf x_T \mid \mathbf x_t)\cdot \nabla_{\mathbf x_t}\ p(\mathbf x_t \mid \mathbf x_0) \right.\\
&\left. +\frac{1}{p(\mathbf x_T \mid \mathbf x_0)}\nabla_{\mathbf x_t} p(\mathbf x_T \mid \mathbf x_t)\cdot \nabla_{\mathbf x_t}\ p(\mathbf x_t \mid \mathbf x_0)+\frac{p(\mathbf x_t \mid \mathbf x_0)}{p(\mathbf x_T \mid \mathbf x_0)}\nabla_{\mathbf x_t} \cdot \nabla_{\mathbf x_t} p(\mathbf x_T\mid \mathbf x_t) \right]\\
&-g_t^2\left[\frac{1}{p(\mathbf x_T \mid \mathbf x_0)}\nabla_{\mathbf x_t} p(\mathbf x_T \mid \mathbf x_t)\cdot \nabla_{\mathbf x_t}\ p(\mathbf x_t \mid \mathbf x_0) + \frac{p(\mathbf x_t \mid \mathbf x_0)}{p(\mathbf x_T \mid \mathbf x_0)}\nabla_{\mathbf x_t} \cdot \nabla_{\mathbf x_t} p(\mathbf x_T\mid \mathbf x_t) \right]\\
=&\frac{1}{2}g_t^2\left[\frac{1}{p(\mathbf x_T \mid \mathbf x_0)}\nabla_{\mathbf x_t}\cdot \left[p(\mathbf x_T \mid \mathbf x_t)\nabla_{\mathbf x_t}p(\mathbf x_t \mid \mathbf x_0) \right] + \frac{1}{p(\mathbf x_T \mid \mathbf x_0)}\nabla_{\mathbf x_t}\cdot \left[p(\mathbf x_t \mid \mathbf x_0)\nabla_{\mathbf x_t}p(\mathbf x_T \mid \mathbf x_t) \right] \right]\\
&-g_t^2 \frac{1}{p(\mathbf x_T \mid \mathbf x_0)}\nabla_{\mathbf x_t}\cdot \left[p(\mathbf x_t \mid \mathbf x_0)\nabla_{\mathbf x_t}p(\mathbf x_T \mid \mathbf x_t) \right]\\
=&\frac{1}{2}g_t^2\left[\nabla_{\mathbf x_t}\cdot \left[p(\mathbf x_t \mid \mathbf x_0, \mathbf x_T)\nabla_{\mathbf x_t}\log p(\mathbf x_t \mid \mathbf x_0) \right] + \nabla_{\mathbf x_t}\cdot \left[p(\mathbf x_t \mid \mathbf x_0, \mathbf x_T)\nabla_{\mathbf x_t}\log p(\mathbf x_T \mid \mathbf x_t) \right] \right]\\
&-g_t^2 \nabla_{\mathbf x_t}\cdot \left[p(\mathbf x_t \mid \mathbf x_0, \mathbf x_T)\nabla_{\mathbf x_t}\log p(\mathbf x_T \mid \mathbf x_t) \right]\\
=&\frac{1}{2}g_t^2\left[\nabla_{\mathbf x_t}\cdot \left[p(\mathbf x_t \mid \mathbf x_0, \mathbf x_T)\nabla_{\mathbf x_t}\log p(\mathbf x_t \mid \mathbf x_0, \mathbf x_T) \right] \right]-g_t^2 \nabla_{\mathbf x_t}\cdot \left[p(\mathbf x_t \mid \mathbf x_0, \mathbf x_T)\nabla_{\mathbf x_t}\log p(\mathbf x_T \mid \mathbf x_t) \right]\\
=&\frac{1}{2}g_t^2 \nabla_{\mathbf x_t} \cdot \nabla_{\mathbf x_t} p(\mathbf x_t \mid \mathbf x_0, \mathbf x_T) -g_t^2 \nabla_{\mathbf x_t}\cdot \left[p(\mathbf x_t \mid \mathbf x_0, \mathbf x_T)\nabla_{\mathbf x_t}\log p(\mathbf x_T \mid \mathbf x_t) \right]
\end{aligned}
\end{equation}
Bring it back to \eqref{51}:
\begin{equation}
\begin{aligned}
\frac{\partial}{\partial t} p(\mathbf x_t \mid \mathbf x_0, \mathbf x_T) 
&=- \nabla_{\mathbf x_t} \cdot \left[\mathbf f(\mathbf x_t, t) p(\mathbf x_t \mid \mathbf x_0, \mathbf x_T) \right] + \frac{1}{2}g_t^2 \nabla_{\mathbf x_t} \cdot \nabla_{\mathbf x_t} p(\mathbf x_t \mid \mathbf x_0, \mathbf x_T) \\
&\quad -g_t^2 \nabla_{\mathbf x_t}\cdot \left[p(\mathbf x_t \mid \mathbf x_0, \mathbf x_T)\nabla_{\mathbf x_t}\log p(\mathbf x_T \mid \mathbf x_t) \right]\\
&=- \nabla_{\mathbf x_t} \cdot \left[[\mathbf f(\mathbf x_t, t) + g_t^2 \nabla_{\mathbf x_t}\log p(\mathbf x_T \mid \mathbf x_t)]p(\mathbf x_t \mid \mathbf x_0, \mathbf x_T) \right] + \frac{1}{2}g_t^2 \nabla_{\mathbf x_t} \cdot \nabla_{\mathbf x_t} p(\mathbf x_t \mid \mathbf x_0, \mathbf x_T)
\end{aligned}
\end{equation}
This is the definition of FP equation of conditional transition probability $p(\mathbf x_t \mid \mathbf x_0, \mathbf x_T)$, which represents the evolution follows the SDE:
\begin{equation}
\mathrm{d}\mathbf{x}_t=\left[\mathbf{f}( \mathbf{x}_t,t) + g^2_t\nabla_{\mathbf x_t} \log p(\mathbf x_T \mid \mathbf x_t) \right]\mathrm{d}t + g_t \mathrm{d}\mathbf{w}_t
\end{equation}
This concludes the proof of the \textbf{Theorem \ref{t2}.}
\section{Experimental Details}\label{Experimental Details}
For all experiments, we use the same noise network, with the network architecture and mainly training parameters consistent with the paper \cite{luo2023image}. This network is similar to a U-Net structure but without group normalization layers and self-attention layers. The steady variance level $\lambda^2$ was set to 30 (over 255), and the sampling step number T was set to 100. In the training process, we set the patch size = 128 \time 128 with batch size = 8 and use Adam \cite{kingma2014adam} optimizer with parameters $\beta_1=0.9$ and $\beta_2=0.99$. The total training steps are 900 thousand with the initial learning rate set to $10^{-4}$, and it decays by half at iterations 300, 500, 600, and 700 thousand. For the setting of $\theta_t$, we employ a flipped version of cosine noise schedule \cite{nichol2021improved}, enabling $\theta_t$ to change from 0 to 1 over time. Notably, to address the issue of $\theta_t$ being too smooth when $t$ closed to 1, we let the coefficient $e^{-\bar \theta_T}$ to be a small enough value $\delta = 0.005$ instead of zero, which represents $\bar \theta_T \approx \sum_{i=0}^{T} \theta_i \mathrm dt = -\log \delta$, as well as $\mathrm dt =-\log \delta/\sum_{i=0}^{T} \theta_i$. Our models are trained on a single 3090 GPU with 24GB memory for about 2.5 days.
\section{Additional Experiments}

\begin{table}[h]
  \centering
  \caption{\textbf{Image Inpainting.} Qualitative comparison with the relevant baselines on CelebA-HQ with thick mask.}
  \vskip 0.15in
  \begin{tabular}{lcccc}
    \toprule
    \textbf{METHOD} & \textbf{PSNR\(\uparrow\)} & \textbf{SSIM\(\uparrow\)} & \textbf{LPIPS\(\downarrow\)} & \textbf{FID\(\downarrow\)} \\
    \midrule
    DDRM&19.48&0.8154&0.1487&26.24\\
    IRSDE&21.12&0.8499&0.1046&11.12\\
    \midrule
    GOUB&\textbf{22.27}&\textbf{0.8754}&\textbf{0.0914}&\textbf{5.64}\\
    \bottomrule
  \end{tabular}
  \vskip -0.1in
\end{table}

\begin{table}[h]
  \centering
  \caption{\textbf{Image Deraining.}  Qualitative comparison with the relevant baselines on Rain100L.}
  \vskip 0.15in
  \begin{tabular}{lcccc}
    \toprule
    \textbf{METHOD} & \textbf{PSNR\(\uparrow\)} & \textbf{SSIM\(\uparrow\)} & \textbf{LPIPS\(\downarrow\)} & \textbf{FID\(\downarrow\)} \\
    \midrule
    PRENET&	37.48&	0.9792&	0.020&	10.9\\
    MAXIM&	38.06&	0.9770&	0.048&	19.0\\
    IRSDE&	38.30&	0.9805&	0.014&	7.94\\
    \midrule
    GOUB&	\textbf{39.79}&	\textbf{0.9830}&	\textbf{0.009}&	\textbf{5.18} \\
    \bottomrule
  \end{tabular}
  \vskip -0.1in
\end{table}

\begin{table}[h]
  \centering
  \caption{\textbf{Image 8$\times$ Super-Resolution.} Qualitative comparison with the relevant baselines on DIV2K.}
  \vskip 0.15in
  \begin{tabular}{lcccc}
    \toprule
    \textbf{METHOD} & \textbf{PSNR\(\uparrow\)} & \textbf{SSIM\(\uparrow\)} & \textbf{LPIPS\(\downarrow\)} & Training Datasets \\
    \midrule
    SRFlow&	23.05&	0.57&	\textbf{0.272}&	DIV2K + Flickr2K\\
    IRSDE&	22.34&	0.55&	0.331&	DIV2K\\
    \midrule
    GOUB&	\textbf{23.17}&	\textbf{0.60}&	0.310&	DIV2K\\
    \bottomrule
  \end{tabular}
\vskip -0.1in
\end{table}

\section{Additional Visual Results}
\begin{figure}[h]
  \centering
  \includegraphics[width=1.0\textwidth]{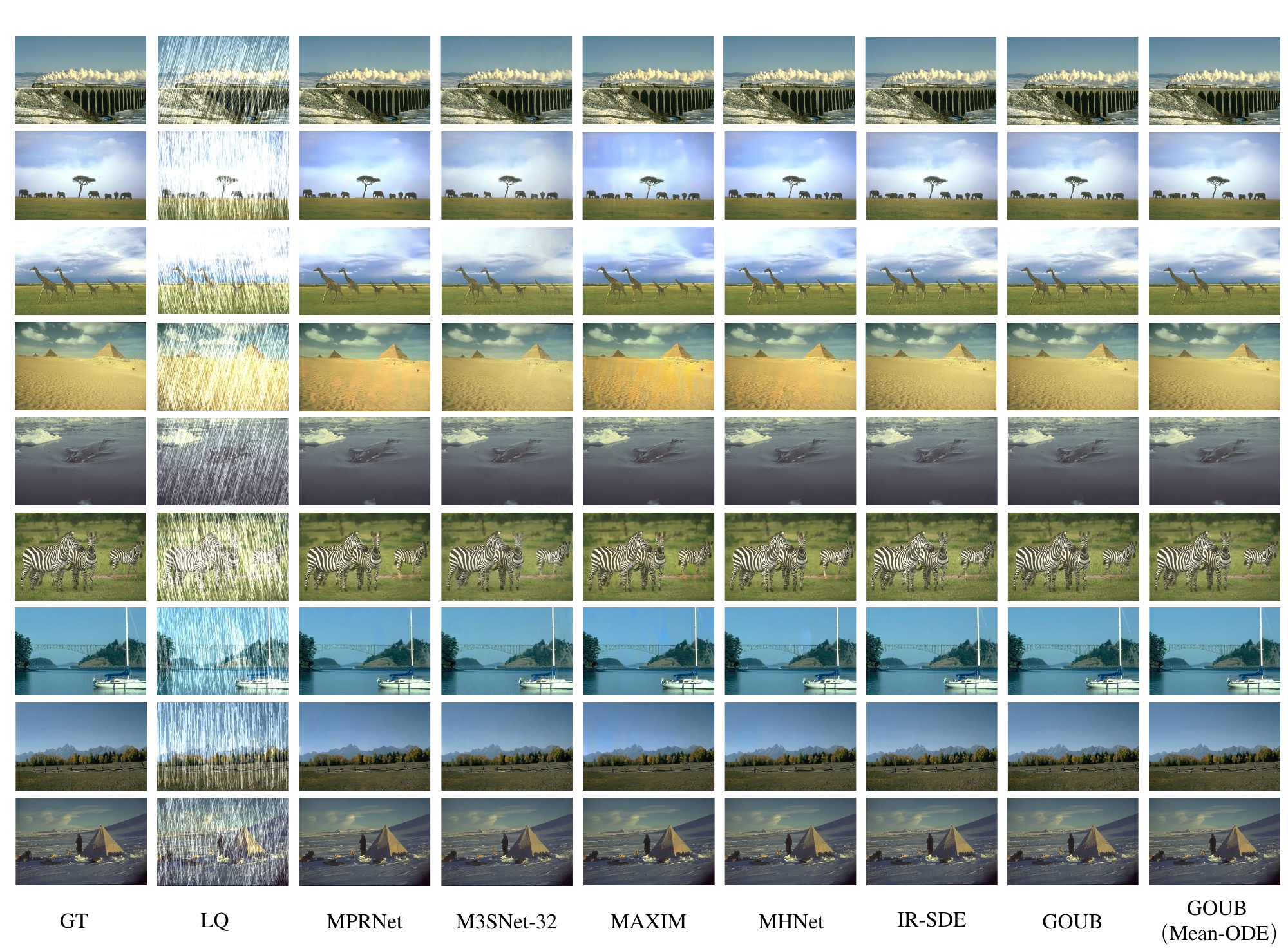}
  \caption{Additional visual results on deraining with Rain100H datasets.}
\end{figure}

\begin{figure}[h]
  \centering
  \includegraphics[width=1.0\textwidth]{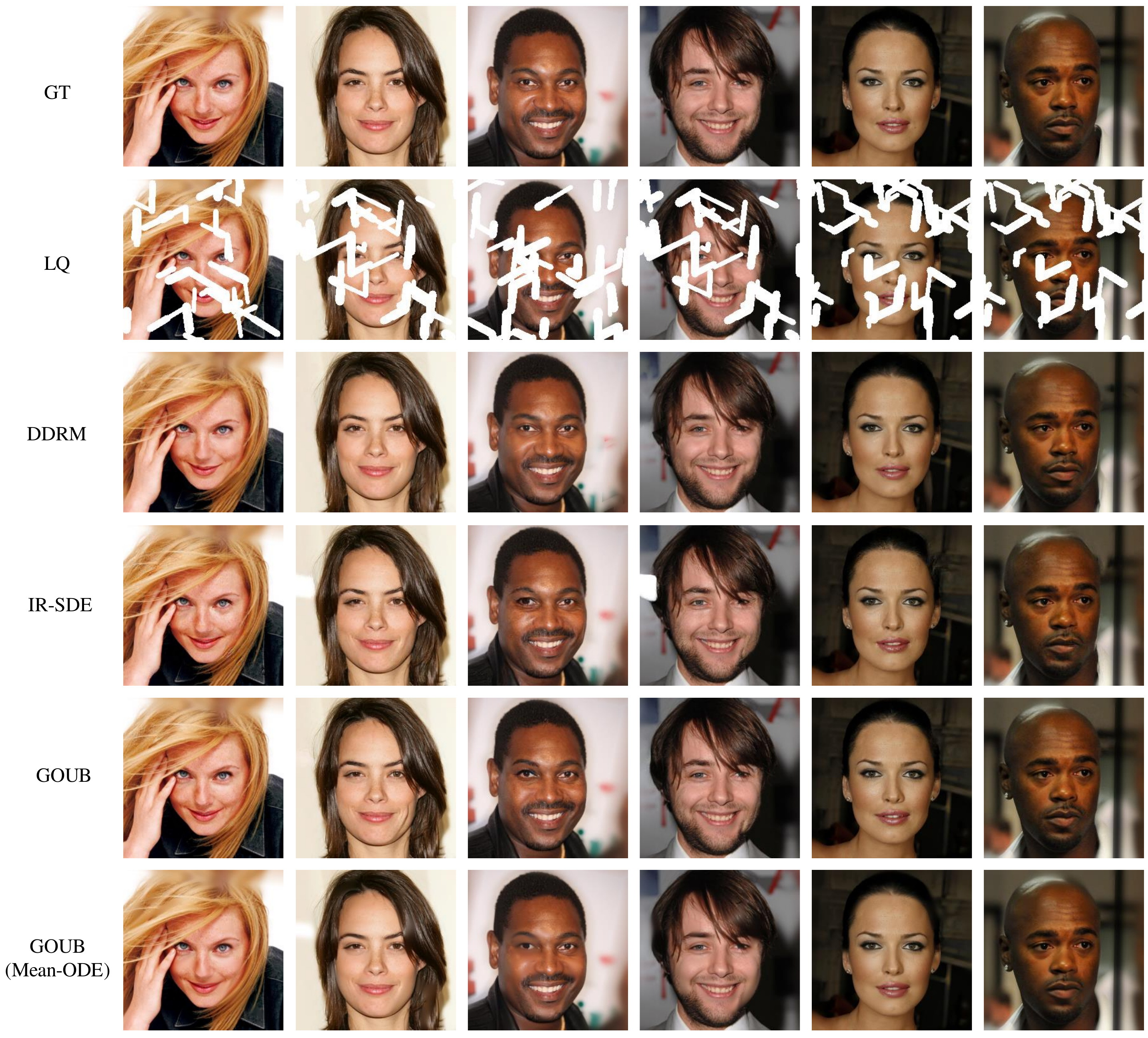}
  \caption{Additional visual results on thin mask inpainting with CelebA-HQ datasets.}
\end{figure}

\begin{figure}[h]
  \centering
  \includegraphics[width=1.0\textwidth]{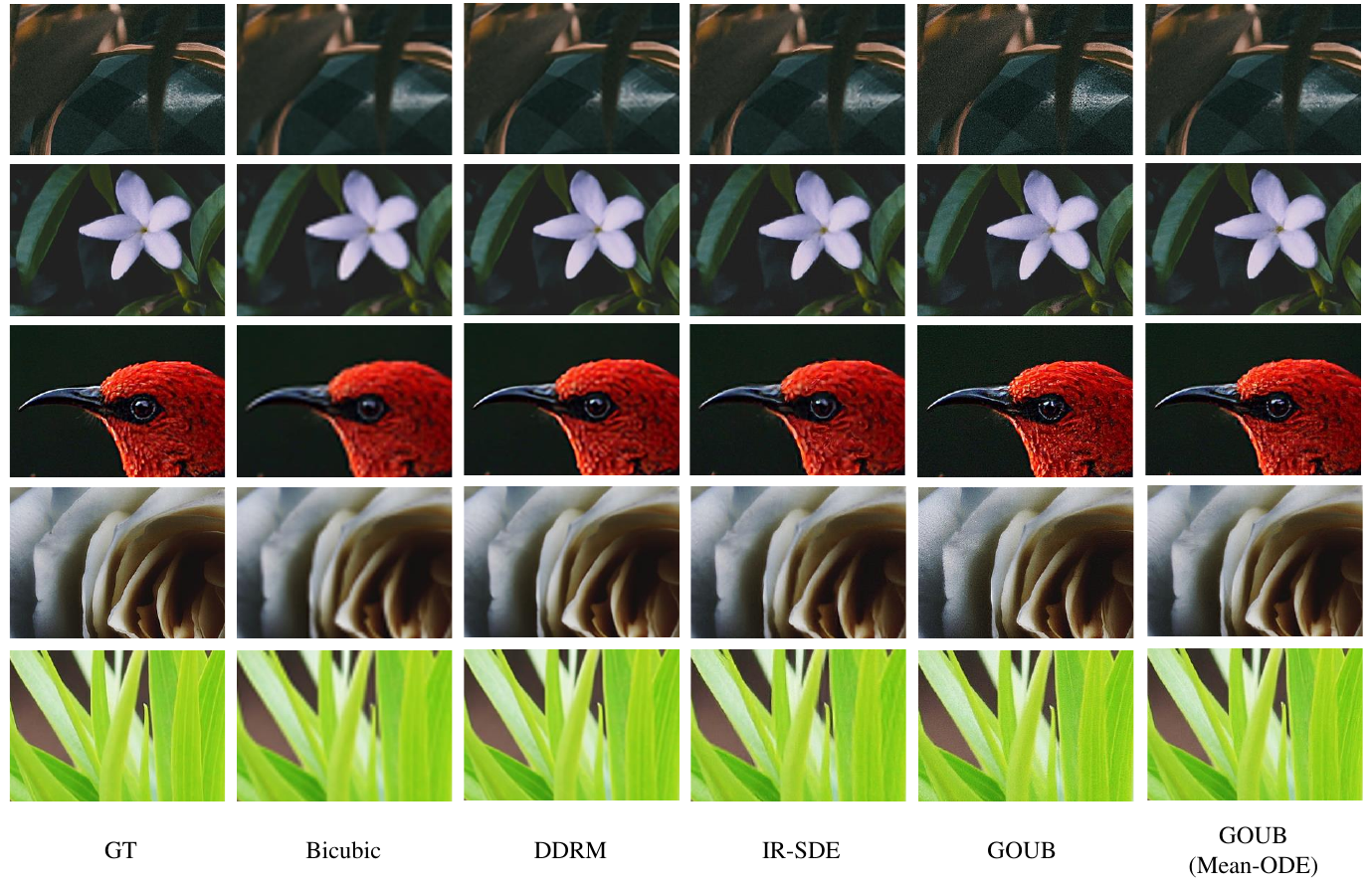}
  \caption{Additional visual results on 4x super-resolution with DIV2K datasets.}
\end{figure}

%%%%%%%%%%%%%%%%%%%%%%%%%%%%%%%%%%%%%%%%%%%%%%%%%%%%%%%%%%%%%%%%%%%%%%%%%%%%%%%
%%%%%%%%%%%%%%%%%%%%%%%%%%%%%%%%%%%%%%%%%%%%%%%%%%%%%%%%%%%%%%%%%%%%%%%%%%%%%%%

\end{document}